\documentclass[10pt,twocolumn,letterpaper]{article}

\usepackage{cvpr}              % To produce the CAMERA-READY version
\usepackage[accsupp]{axessibility}  % Improves PDF readability for those with disabilities. 

\usepackage{amsmath}
\usepackage{algorithm}
\usepackage{algpseudocode}
\usepackage{multirow}

\definecolor{cvprblue}{rgb}{0.21,0.49,0.74}
\usepackage[pagebackref,breaklinks,colorlinks,allcolors=cvprblue]{hyperref}
\usepackage{parskip}

\def\paperID{} % *** Enter the Paper ID here
\def\confName{CVPR}
\def\confYear{2026}

\title{Beyond Semantics: Disentangling Information Scope in Sparse Autoencoders for CLIP}

\author{
Yusung Ro$^{1}$ \qquad
Jaehyun Choi$^{1,2}$ \qquad
Junmo Kim$^{1}$\\[0.5em]
$^{1}$KAIST \qquad
$^{2}$Korea AI Safety Institute, ETRI\\[0.5em]
{\tt\small ysro3067@kaist.ac.kr, pre6ent@etri.re.kr, junmo.kim@kaist.ac.kr}
}

\begin{document}
\maketitle
\setlength{\parskip}{0.3em}

\begin{figure*}[htbp]
    \centering
    % 0.8\linewidth 또는 0.9\linewidth 등 원하는 비율로 조절
    % 1.0\linewidth로 꽉 채우면 답답해 보일 수 있으니 0.8~0.9가 적절
    \includegraphics[width=0.9\linewidth]{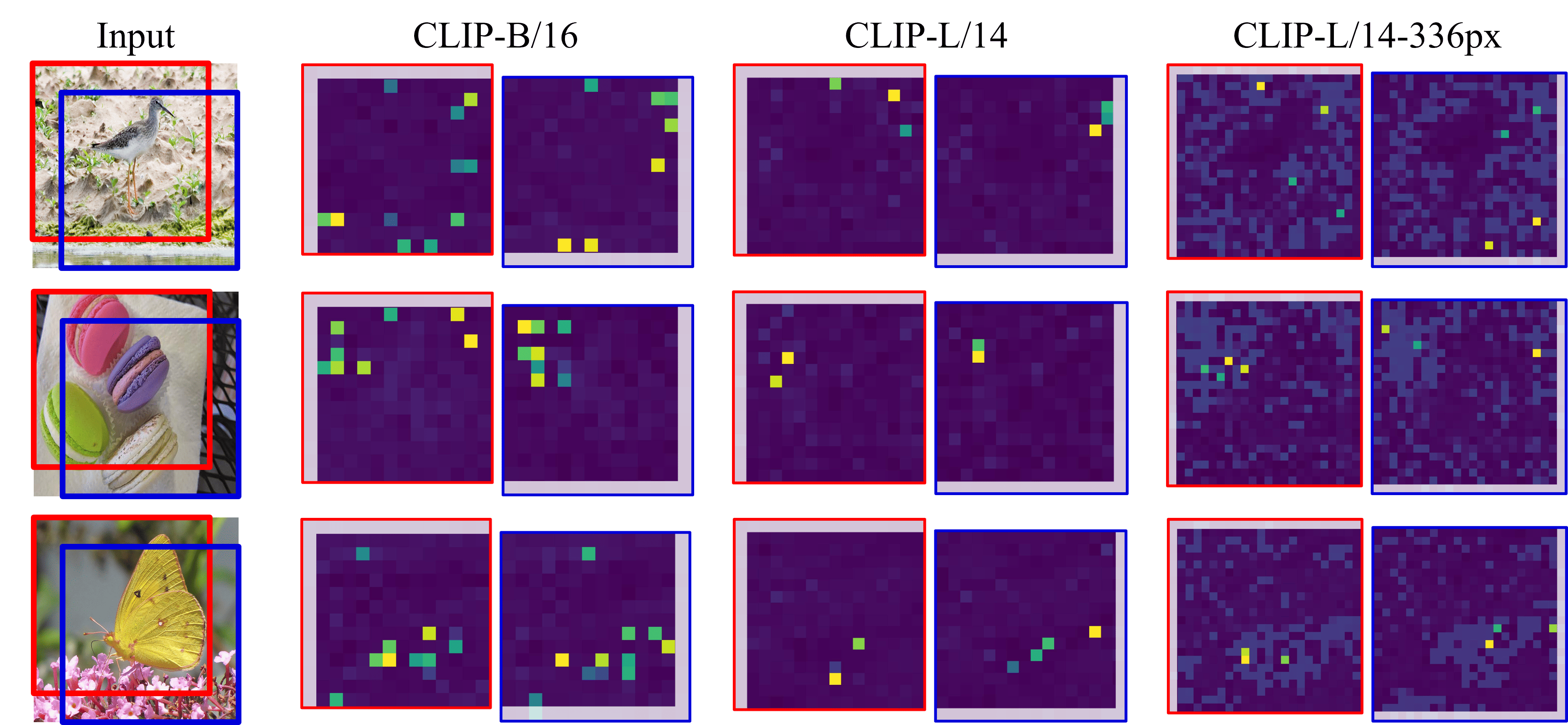} 
    
    % \caption은 figure*에 직접 답니다.
    \caption{\textbf{Spatial instability of CLIP outlier tokens under Shifted Context Cropping (SCC).} We analyze the spatial consistency of patch-token L2 norms using two overlapping crops (red: top-left; blue: bottom-right) translated by a single patch. The heatmaps visualize patch-token L2 norms for each crop across different CLIP ViT models. The green and yellow spots, indicating outlier tokens, do not maintain consistent spatial locations between the two overlapping crops despite this minimal contextual shift. This shows that outlier token locations are highly unstable under SCC.}
    \label{fig:pcp_outlier} 
    % \vspace{-0.5cm}
\end{figure*}

\begin{abstract}
Sparse Autoencoders (SAEs) have emerged as a powerful tool for interpreting the internal representations of CLIP vision encoders, yet existing analyses largely focus on the semantic meaning of individual features. We introduce information scope as a complementary dimension of interpretability that characterizes how broadly an SAE feature aggregates visual evidence, ranging from localized, patch-specific cues to global, image-level signals.
We observe that some SAE features respond consistently across spatial perturbations, while others shift unpredictably with minor input changes, indicating a fundamental distinction in their underlying scope. To quantify this, we propose the Contextual Dependency Score (CDS), which separates positionally stable local scope features from positionally variant global scope features.
Our experiments show that features of different information scopes exert systematically different influences on CLIP’s predictions and confidence. These findings establish information scope as a critical new axis for understanding CLIP representations and provide a deeper diagnostic view of SAE-derived features.
\end{abstract}

\section{Introduction}

Sparse Autoencoders (SAEs) have recently become a central tool for analyzing internal representations in vision and language models \cite{gao2024scaling, cunningham2023sparse, bricken2023monosemanticity, park2025decoding, elhage2022toy}. Their appeal lies in their ability to decompose dense, polysemantic \cite{olah2017feature} representations into a linear combination of sparse, monosemantic representations. 
Motivated by their success in large language models, several works have applied SAEs to CLIP \cite{han2025causal, zaigrajew2025interpreting, daujotas2024case, daujotas2024interpreting, rao2024discover} and Vision Transformers (ViTs) \cite{  thasarathan2025universal, stevens2025sparse, hindupur2025projecting, baccouche2012spatio}, uncovering meaningful patch-level visual concepts and enabling new forms of feature-level analysis.

Despite this progress, SAE-based interpretability research focuses mainly on what each feature represents semantically. 
This concept-centric approach, whether based on manual visual inspection or automated labeling pipelines \cite{zhang2025large, olson2025probing}, inherently struggles to determine what kind of concept actually drives a feature's activation.
% For instance, it remains ambiguous whether a feature genuinely captures a holistic concept (e.g., a "dog") or merely reacts to a localized texture (e.g., "fur").
For example, a feature labeled as ``dog" may either reflect evidence integrated across the whole object or respond mainly to a local texture such as fur.
This limitation becomes more severe as SAE dictionaries grow and features become increasingly fine-grained, making semantic interpretation alone too brittle to characterize their functional roles consistently.

Motivated by this limitation, we shift our focus from semantic identity to information scope. 
Rather than asking only which concept a feature represents, we ask over what spatial extent it aggregates evidence. 
Some SAE features rely on localized cues, whereas others integrate broader image context. 
We refer to this property as a feature’s information scope and quantify it through its contextual dependency.
This dimension is essential for understanding the functional roles that different features play in a model’s predictions.

We further ground this perspective by examining outlier tokens in Vision Transformers \cite{darcet2023vision, jiang2025vision}, a small subset of patch tokens with unusually high norms that have been associated with global information aggregation.
Under Shifted Context Cropping (SCC), where overlapping crops are translated by a single patch stride, these tokens exhibit striking spatial instability (Figure \ref{fig:pcp_outlier}). 
Their positions shift unpredictably even under such minimal contextual shifts.
This behavior suggests an important asymmetry between local and global signals. 
Signals associated with global context are highly sensitive to contextual shifts, whereas localized signals remain more stably anchored to visual content. 
We take this contrast between local stability and global instability as the key intuition underlying our metric.

To quantify this property, we introduce the Contextual Dependency Score (CDS), a metric that measures how a feature’s spatial activation changes across small contextual shifts.
Low-CDS features remain spatially stable across contextual shifts and correspond to local-scope information, whereas high-CDS features vary more across such shifts and correspond to global-scope information.
CDS enables a principled partition of SAE features into low-CDS and high-CDS groups, corresponding to local and global information scope, respectively, and reveals structure that cannot be captured through semantic analysis alone.

We validate this partition through feature-group removal experiments using linear probing on image classification, semantic segmentation, and monocular depth estimation. 
By selectively removing either low-CDS or high-CDS features from CLIP embeddings via the SAE, we measure their distinct functional roles across tasks. 
Our results demonstrate that low-CDS features primarily carry fine-grained spatial cues, whereas high-CDS features convey global scene structure and image-level context.

In summary, our contributions are:
\begin{enumerate}
    \item We introduce contextual dependency as a new interpretability dimension for SAE features, capturing their information scope as the spatial extent over which they aggregate evidence.
    \item We propose the Contextual Dependency Score (CDS), a metric that quantifies contextual dependency through changes in a feature’s spatial activation across small contextual shifts.
    \item We show that CDS induces a principled partition of SAE features into low-CDS and high-CDS groups, corresponding to local and global information scope, and analyze their functional roles across downstream tasks.
\end{enumerate}

\section{Related Work}
\label{sec:related_work}

\subsection{Sparse Autoencoders for Vision Transformers}
\noindent
Research analyzing ViTs \cite{dosovitskiy2020image} using SAEs has primarily focused on CLIP \cite{radford2021learning}. 
Applying SAEs to pooled CLIP embeddings has yielded interpretability results, such as extracting concepts \cite{fry2024towards} used for tasks like diffusion model steering \cite{daujotas2024case, daujotas2024interpreting} and forming concept bottlenecks \cite{rao2024discover}. 
While these studies focus on decomposing aggregated, CLIP embeddings, another line of research has begun to apply SAEs directly to the more granular visual patch tokens within the transformer's layers. 
For example, PatchSAE \cite{lim2024sparse} classifies features learned from visual tokens by their class- and dataset-level activations and evaluates the influence on classification via feature ablation. 
To further analyze these features, \cite{pach2025sparse} introduce a monosemanticity evaluation framework and the Monosemanticity Score (MS), which quantifies how well individual SAE features correspond to monosemantic concepts. 
Additionally, USAE \cite{thasarathan2025universal} introduced a framework to jointly train SAEs across multiple vision models, facilitating the analysis of commonly learned features. 
These prior works commonly approach the analysis of SAE features from a semantic perspective, aiming to identify what concepts the features represent. 
In contrast, our work introduces a different axis of analysis. 
We investigate the information scope of SAE features, distinguishing between those that encode localized information and those that capture broder global context.

\subsection{Outlier Tokens in Vision Transformers}
\noindent
The scaling of ViTs to large data and model sizes has led to the emergence of outlier tokens \cite{darcet2023vision, jiang2025vision, bachregisters}. 
This phenomenon describes patch tokens that acquire exceptionally high norms during both training and inference and receive strong attention from the CLS token.
A systematic analysis of this phenomenon was first presented in \cite{darcet2023vision}, which found that these high-norm tokens emerge primarily in redundant, low-information patches, such as background areas. 
Further analysis revealed that these tokens lose significant local information. 
Evidence for this was provided through probing tasks where high-norm tokens performed poorly on tasks requiring local detail, such as position prediction and pixel reconstruction. 
Conversely, despite this loss of local detail, the same tokens were found to condense global, image-level information. 
This was demonstrated by their superior performance when used as a sole global representation for image classification \cite{darcet2023vision}, indicating they had captured global information.
Overall, these studies describe outlier tokens as a structural behavior of large ViTs, where token abandon local, patch-level detail to encode global, image-level information.
Our work applies SAEs to disentangle these outlier tokens, aiming to identify the specific features that correspond to global information.
\section{Preliminary}
\subsection{Sparse Autoencoders}
SAEs are an unsupervised model class designed for sparse dictionary learning \cite{olshausen1997sparse}. The goal is to learn an overcomplete set of features that can linearly reconstruct a high-dimensional signal in a sparse manner. In the context of model interpretability, SAEs are trained on the internal embedding vectors $v \in \mathbb{R}^{d}$ from a specific layer of a pre-trained model, such as a ViT.

\noindent
An SAE consists of two linear layers: an encoder $W_{enc}\in \mathbb{R}^{d\times q}$ for projecting inputs into the sparse feature space and a decoder $W_{dec}\in \mathbb{R}^{q\times d}$ for reconstruction. 
Both layers share a bias term $b\in\mathbb{R}^d$. 
The dictionary size $M$ is defined by an expansion factor $\epsilon>1$ as $q:=d\times \epsilon$.

BatchTopK SAE \cite{bussmann2024batchtopk} replaces the continuous sparsity regularization (e.g., L1 penalty) with an explicit TopK activation constraint. The encoder and decoder are defined as: 
\begin{equation}
\phi(v) = \text{BatchTopK}(W_\text{enc}^T (v - b)), \quad \hat{v} = W_\text{dec}^T\phi(v) + b
\end{equation}
where $\text{BatchTopK}(\cdot)$ retains only the top $n\times k$ activations per batch of $n$ samples and sets all others to zero.
This hard sparsity constraint enforces a fixed sparsity ratio, leading to more stable and interpretable features.
The reconstruction objective is:
\begin{equation}
\mathcal{L}_{recon} = ||v - \hat{v}||_2^2.    
\end{equation}
During training, some latent features may become inactive dead features, never firing across the dataset. To re-activate these dead features and maintain a balanced dictionary, an auxiliary reconstruction loss \cite{gao2024scaling} is introduced.
For each training batch, features that have not fired for more than a threshold number of steps are marked as dead.
Among them, the top-$K_{aux}$ features with the highest potential activation values are temporarily reactivated and employed to reconstruct the model's reconstruction error, $e = v - \hat{v}$.
The auxiliary loss is then defined as:
\begin{equation}
\mathcal{L}_{\text{aux}} = ||e - \hat{e}||_2^2,
\quad
\hat{e} = W_{\text{dec}}^T \, \phi_{\text{aux}}(v),
\end{equation}
where $\phi_{aux}(v)$ contains only the activations of the selected dead features. Finally, the overall training objective combines the main and auxiliary losses: 
\begin{equation}
\mathcal{L}_{total} = \mathcal{L}_{recon} + \alpha \mathcal{L}_{aux},
\end{equation}
where $\alpha$ controls the contribution of the auxiliary term.
This auxiliary objective prevents feature collapse, encourages uniform feature utilization, and promotes a more diverse and expressive dictionary of learned features.

\subsection{ViT Notation}
We first define the standard ViT tokenization process to establish our notation. 
A given input image $I_{\text{orig}}$ is resized to a fixed resolution, yielding $I_{\text{resized}} \in \mathbb{R}^{H \times W \times C}$.
Following the standard ViT architecture, this resized image $I_{resized}$ is then divided into $p \times p$ patches of size $n \times n$ pixels, where $H = p \cdot n$ and $W = p \cdot n$. 
The ViT encoder then processes these patches, producing a sequence of patch tokens $V = [v_0, \dots, v_{N-1}] \in \mathbb{R}^{N \times d}$, where $N=p^2$ is the total number of patch tokens and $d$ is the embedding dimension.

% In line with observations from prior work \cite{darcet2023vision}, we identify a small subset of these patch tokens that exhibit exceptionally high L2 norms. We refer to these as outlier tokens, which are hypothesized to aggregate abstract, global information rather than the local content of their corresponding patch. Because the distribution of these norms varies significantly across different models, we define an outlier token based on a model-specific absolute threshold, $\tau$. Specifically, for a given patch token $e_i$, we compute its L2 norm $L_i = ||e_i||_2$. The set of outlier indices $S_{out}$ is defined as:
% \begin{equation}
% S_{out} = \{i \mid L_i > \tau \}.
% \end{equation}
% Based on empirical observation of the norm distributions for the models used in this study, we set $\tau=60$ for CLIP-ViT-base-patch16 and $\tau=100$ for both CLIP-ViT-large-patch14 and CLIP-ViT-large-patch14-336px \cite{radford2021learning}.
\begin{figure*}[htbp]
    \centering
    % 0.8\linewidth 또는 0.9\linewidth 등 원하는 비율로 조절
    % 1.0\linewidth로 꽉 채우면 답답해 보일 수 있으니 0.8~0.9가 적절
    \includegraphics[width=0.9\linewidth]{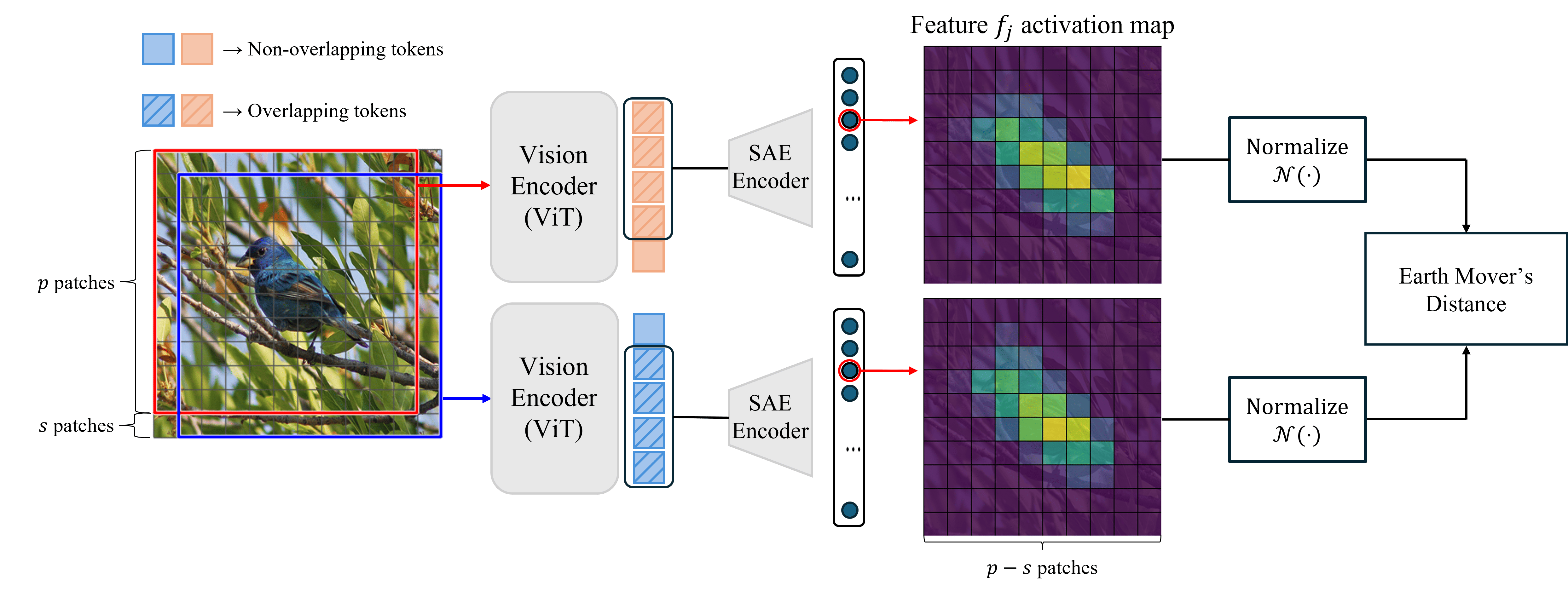} 
    
    % \caption은 figure*에 직접 답니다.
    \caption{\textbf{Framework for quantifying our Contextual Dependency Score (CDS) of SAE features.} We evaluate the contextual dependency of SAE features using CDS. This score is derived from the Earth Mover's Distance (EMD) computed between the normalized activation maps of two shifted input crops (red and blue) as shown for a single pair. The final CDS for a feature is the average of these EMD scores over $k_{CDS}$ representative images. We process only the embeddings from the overlapping region.}
    \label{fig:perturbation_consistency} % 레이블도 의미있게 변경
    \vspace{-0.5cm}
\end{figure*}

\section{Method}

\textbf{Outlier token definition.}
In Vision Transformers, certain patch tokens exhibit exceptionally large embedding norms and consistently emerge across diverse inputs. 
These tokens, often referred to as outlier tokens, have been associated with global information aggregation rather than strictly local patch content.
Following prior work \cite{darcet2023vision}, we define outlier tokens as patch tokens with norms exceeding a predefined threshold $\tau$. 
For a patch-token sequence $V = [v_0, \dots, v_{N-1}] \in \mathbb{R}^{N \times d}$, we define the outlier binary mask $S_{\text{out}}(V)\in\{0,1\}^N$ element-wise as
\begin{equation}
\bigl(S_{\text{out}}(V)\bigr)_a =
\begin{cases}
1 & \text{if } \|v_a\|_2 > \tau,\\
0 & \text{otherwise},
\end{cases}
\quad a = 0,\dots,N-1.
\end{equation}

\subsection{Shifted Context Crop}

\begin{algorithm}[t]
\caption{\textbf{Shifted Context Crop (SCC)}}
\label{alg:pcp}

\begin{algorithmic}[1]

\Require Original image $I_{\text{orig}}$, grid per image $p \times p$, pixel per patch $n\times n$, shifting factor $s$
\Ensure Two overlapping shifted crops $I_1$, $I_2$

% --- STEP 1 ---
\State \textbf{Form expanded image}
\State $I_{\text{resized}} \gets I_{\text{orig}}$ resized to $(p+s)n \times (p+s)n$
\Comment{adds $s$-patch border on all sides}

% --- STEP 2 ---
\State \textbf{Define crop operator}
\State $\text{Crop}(I,(x,y),(w,h))$: return sub-image of $I$ starting at pixel $(x,y)$ with width $w$ and height $h$

% --- STEP 3 ---
\State \textbf{Generate shifted crop images}
\State $I_1 \gets \text{Crop}(I_{\text{resized}},\ (0,0),\ (pn,pn))$
\State $I_2 \gets \text{Crop}(I_{\text{resized}},\ (sn,sn),\ (pn,pn))$

% % --- STEP 4 ---
% \State \textbf{Patch correspondences}
% \For{each $(r,c)$ in $[0,\,p-s-1] \times [0,\,p-s-1]$}
%     \State $P_2(r,c) \equiv P_1(r+s,\ c+s)$
%     \Comment{pixel-identical overlapping patches}
% \EndFor

\State \Return $I_1, I_2$

\end{algorithmic}
\label{alg:PCP}
\end{algorithm}

To investigate the spatial stability of outlier tokens, we design a controlled experiment under the hypothesis that a token's outlier status is not intrinsic to its patch content, but instead depends strongly on broader context, such as surrounding patches and its absolute position in the image.

\noindent
Our Shifted Context Crop (SCC) method, formally defined in Algorithm \ref{alg:PCP}, generates two images, $I_1$ and $I_2$, from a single image $I_{\text{orig}}$.
This cropping strategy results in $I_2$ being a spatial shift of $I_1$ by exactly $sn$ pixels horizontally and vertically. 
% Consequently, a large overlap exists where the patch content of $P_2(r, c)$ (the patch at grid row $r$, col $c$ in $I_2$) is pixel-for-pixel identical to the patch content of $P_1(r+s, c+s)$ (the patch at grid row $r+s$, col $c+s$ in $I_1$), for all $r, c \in [0, \dots, p-s-1]$, creating an overlapping grid of $(p-s) \times (p-s)$ identical patches. 
Consequently, a large overlap exists between the two crops.
For all $r, c \in [0, \dots, p-s-1]$, the patch content of $P_2(r, c)$ (the patch at grid row $r$, col $c$ in $I_2$) is pixel-identical to that of $P_1(r+s, c+s)$ (the patch at grid row $r+s$, col $c+s$ in $I_1$), forming an overlapping grid of $(p-s) \times (p-s)$ identical patches. 
For this shared set of patches, the applied shift isolates purely contextual factors: (1) the absolute positional embedding applied to patch $P_2(r, c)$ is different from that applied to $P_1(r+s, c+s)$, and (2) the set of non-overlapping patches that form the context in the self-attention mechanism is different for each image.

\noindent
\subsection{Analyzing outlier tokens via SCC} 
\label{sec:PCP}
To quantify the instability of outlier tokens, we utilize the proposed SCC method. 
The two generated images, $I_1$ and $I_2$, are independently fed into the ViT. 
We focus our analysis on a specific transformer layer $L$. 
For each image $i \in \{1,2\}$, let $V_i^{(L-1)} = [v_{i,0}^{(L-1)}, \dots, v_{i,N-1}^{(L-1)}] \in \mathbb{R}^{N \times d}$ denote the patch-token sequence at the output of the $(L-1)$-th layer, which serves as the input to layer $L$. We then define the corresponding outlier mask as
\[
S_i := S_{\text{out}}(V_i^{(L-1)}) \in \{0,1\}^N.
\]
From layer $L$, we extract the CLS-to-patch attention vectors for all heads, denoted by $\{A_{i,h} \in \mathbb{R}^{N}\}_{h=1}^{H}$, where $H$ is the number of attention heads.
We restrict the analysis to the $(p-s)\times(p-s)$ overlapping patch region shared by $I_1$ and $I_2$. Let $\Omega(\cdot)$ denote this restriction operator. 
We define
\[
S_i' := \Omega(S_i), \qquad A_{i,h}' := \Omega(A_{i,h}),
\]
where the prime symbol denotes restriction to the overlapping region. 
Thus, $S_i' \in \{0,1\}^{(p-s)\times(p-s)}$ and $A_{i,h}' \in \mathbb{R}^{(p-s)\times (p-s)}$.
For each head $h$, we use the restricted outlier masks to separate the attention vectors into non-outlier and outlier components. 
The non-outlier attention is defined as
\begin{equation}
A_{\text{non}, h, i} = A'_{i,h} \odot (1-S'_i),
\end{equation}
and the outlier attention is defined as
\begin{equation}
A_{\text{out}, h, i} = A'_{i,h} \odot S'_i,
\end{equation}
where $\odot$ denotes element-wise multiplication.

Finally, for each head $h$, we measure the dissimilarity between the corresponding attention maps from $I_1$ and $I_2$ using the Earth Mover's Distance (EMD) \cite{rubner2000earth}. 
Each restricted attention vector is interpreted as a distribution over the $(p-s)\times(p-s)$ overlapping patch grid, and the transport cost in EMD is computed using the Euclidean distance between 2D patch coordinates on this grid.
Each map is first normalized into a probability distribution. 
Let $\mathcal{N}(\cdot)$ denote the normalization operator, e.g., $\mathcal{N}(A)=A/\sum A$. We then compute
\begin{equation}
\begin{split}
D_{\text{non}, h} &= \mathrm{EMD}\bigl(\mathcal{N}(A_{\text{non}, h, 1}), \mathcal{N}(A_{\text{non}, h, 2})\bigr), \\
D_{\text{out}, h} &= \mathrm{EMD}\bigl(\mathcal{N}(A_{\text{out}, h, 1}), \mathcal{N}(A_{\text{out}, h, 2})\bigr).
\end{split}
\end{equation}
To compare the overall instability of outlier and non-outlier tokens, we average these scores across all $H$ heads:
\begin{equation}
\begin{split}
\bar{D}_{\text{non}} &= \frac{1}{H} \sum_{h=1}^{H} D_{\text{non}, h}, \\
\bar{D}_{\text{out}} &= \frac{1}{H} \sum_{h=1}^{H} D_{\text{out}, h}.
\end{split}
\end{equation}
The resulting $\bar{D}_{\text{non}}$ and $\bar{D}_{\text{out}}$ quantify the average contextual instability of non-outlier and outlier attention patterns, respectively. Higher EMD values indicate greater instability under contextual shifts.

\subsection{Contextual Dependency Score}
\label{sec:CIS}
The analysis framework proposed in Section \ref{sec:PCP} allows for a binary comparison of stability between outlier and non-outlier tokens. 
However, this approach does not capture the full spectrum of behavior, nor does it quantify the stability of individual features that may contribute to these tokens. 
To move beyond this binary classification and measure the contextual stability of each individual feature in our trained SAE dictionary, we propose a new metric: the Contextual Dependency Score (CDS). 
This metric allows us to move beyond a simple outlier vs. non-outlier analysis and measure the stability of all features on a continuous spectrum.

\noindent
The CDS for a specific SAE feature $f_j$ (with $j \in {1, \dots, q}$) is calculated by the average dissimilarity of the activation maps across image pairs obtained by our SCC method. 
This process, illustrated in Figure \ref{fig:perturbation_consistency}, involves four steps:

\noindent
1. \textbf{Representative Image Selection}: For each SAE feature $f_j$, we first identify its $\text{top}-k_{CDS}$ most strongly activating images from an image dataset. This ensures that the analysis is performed on images where the feature is salient.

\noindent
2. \textbf{Shifted Context Crop}: For each of the $k_{CDS}$ images, we apply SCC with the shifting factor $s$, to generate two images, $I_1$ and $I_2$. These images share a $(p-s) \times (p-s)$ overlapping patch grid with pixel-for-pixel identical content but differ in their context.

\noindent
3. \textbf{Per-Feature Activation Map Extraction}:
We pass both $I_1$ and $I_2$ through the ViT to obtain their patch embeddings, extracting only the tokens corresponding to the shared $(p-s) \times (p-s)$ overlapping grid. 
These aligned embeddings are then passed through the trained SAE encoder. 
From the resulting dictionary activations, we record the scalar activation of feature $f_j$ at each spatial location.
For the $m$-th representative image, this yields two activation maps: $M_{j,1}^{(m)}$ and $M_{j,2}^{(m)} \in \mathbb{R}^{(p-s) \times (p-s)}$.

\noindent
4. \textbf{Normalized EMD Calculation}: 
We first normalize $M_{j,1}^{(m)}$ and $M_{j,2}^{(m)}$ to treat them as 2D spatial probability distributions. 
We then compute the EMD between them on the overlapping patch grid, using the Euclidean distance between 2D patch coordinates as the ground transport cost. 
Crucially, different ViT architectures employ varying spatial resolutions for their patches.
Since the raw EMD scales linearly with the spatial dimension, direct cross-model comparisons are invalid.
To ensure an architecture-agnostic metric, we normalize the EMD by the diagonal distance $D_{grid}=(p-s) \sqrt{2}$.
The final CDS is obtained by averaging these EMD values over the $k_{CDS}$ images:
\begin{equation} \label{eq:CIS}
\text{CDS}_j = \frac{1}{k_{CDS} \cdot D_{grid}} \sum_{m=1}^{k_{CDS}} \text{EMD}(\mathcal{N}(M_{j,1}^{(m)}), \mathcal{N}(M_{j,2}^{(m)})).
\end{equation}

\noindent
The CDS quantifies the robustness of a feature's activation against changes in context, independent of content. 
A low CDS indicates that a feature is tied more closely to local content, whereas a high CDS indicates stronger dependence on broader context.
\begin{figure*}[t]
    \centering
    \subfloat[CLIP-B/16 \label{fig:sub1}]{
        \includegraphics[width=0.32\textwidth]{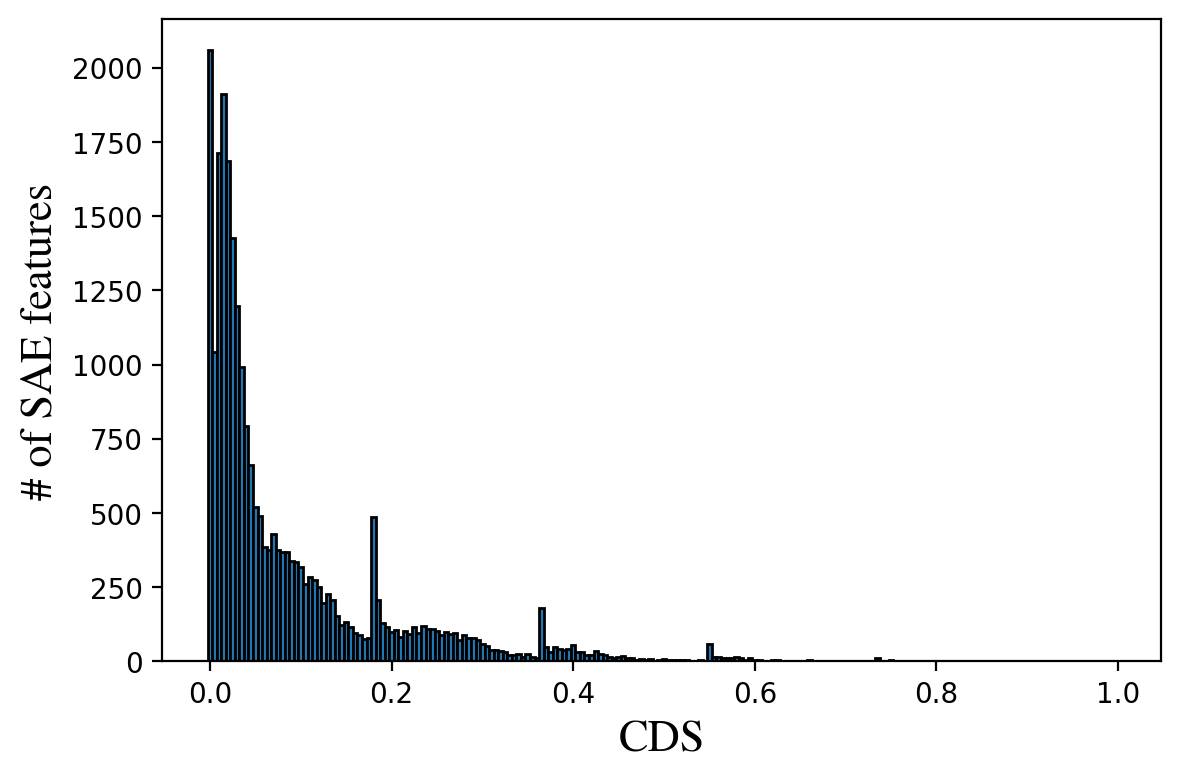}
    }
    \hfill
    \subfloat[CLIP-L/14 \label{fig:sub2}]{
        \includegraphics[width=0.32\textwidth]{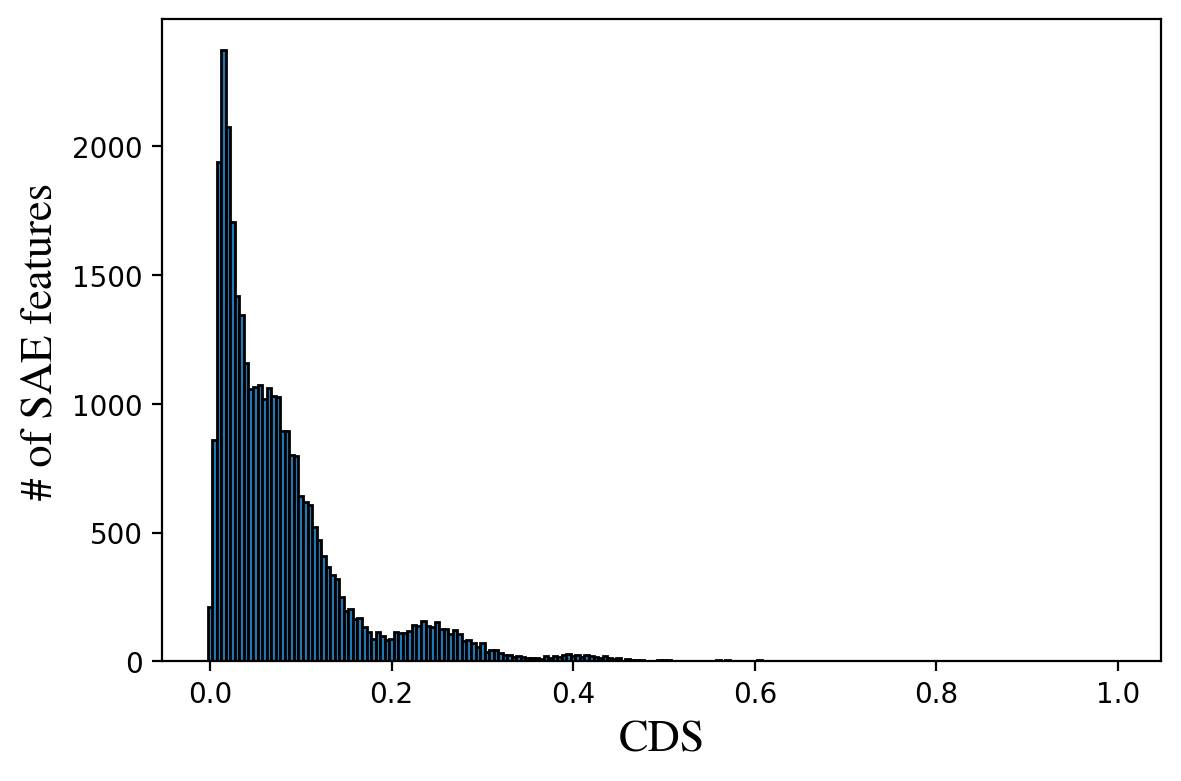}
    }
    \hfill
    \subfloat[CLIP-L/14-336px \label{fig:sub3}]{
        \includegraphics[width=0.32\textwidth]{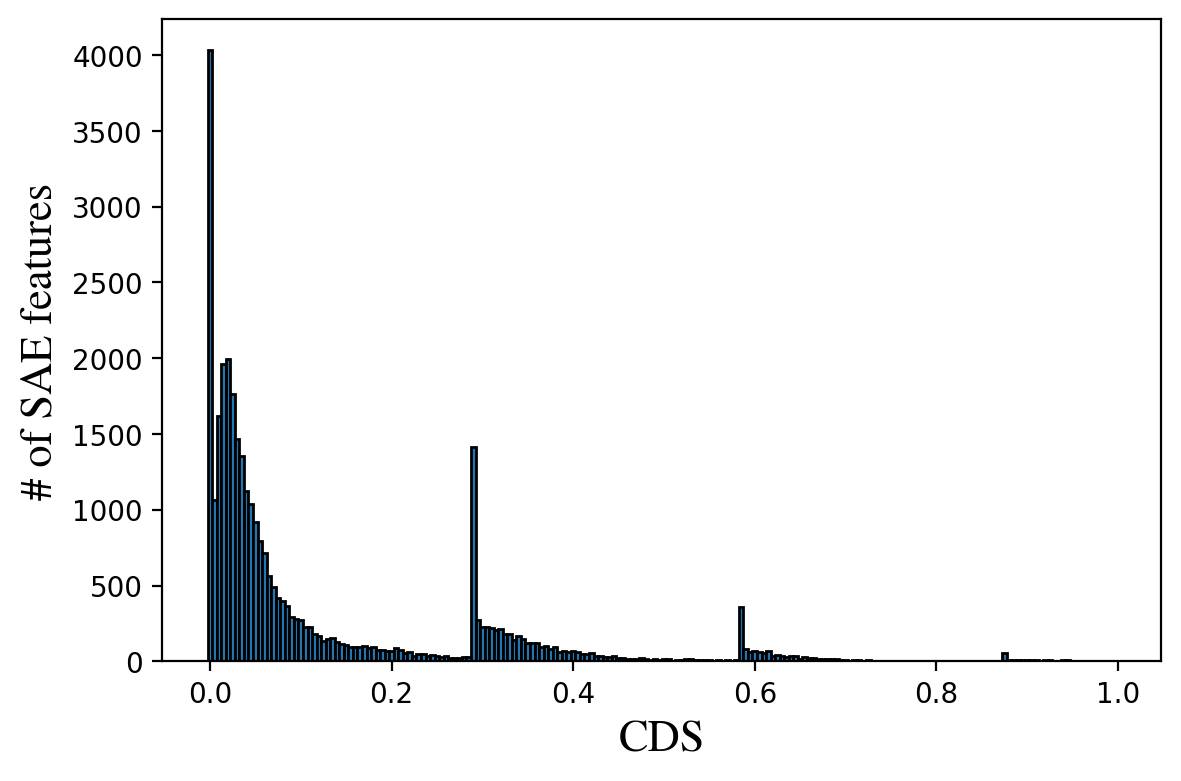}
    }
    \vspace{-0.3cm}
    \caption{\textbf{Distribution of Contextual Dependency Score (CDS).} Histogram of CDS values for SAE features across three CLIP models, showing a dominant low-CDS peak together with distinct peaks in the high-CDS region. This distribution motivates partitioning SAE features into low-CDS and high-CDS groups.}
    \label{fig:cis_result}
    \vspace{-0.5cm}
\end{figure*}

% \begin{figure*}[htbp] % 만약 2단이 아니라 전체 페이지 너비로 그림을 넣고 싶다면 figure* 사용

\section{Experiments}

\subsection{Experimental Settings}
\label{sec:experimental_setting}

To analyze feature properties in CLIP vision encoders, we trained separate BatchTopK SAEs for three ViT-based CLIP vision encoders \cite{radford2021learning}: B/16, L/14, and L/14-336px. 
For all models, we set the expansion factor to $\epsilon=32$ and used $K=30$ in BatchTopK. 
% The resulting SAE dictionaries contain 24,576 features for CLIP-ViT-B/16 and 32,768 features for CLIP-ViT-L/14 and CLIP-ViT-L/14-336px. 
All SAEs were trained on ImageNet-1K \cite{deng2009imagenet} training set. 
For each SAE, we randomly sampled four visual patch tokens per image from the final transformer layer of the corresponding CLIP model. 
All subsequent analyses are conducted on these layers. 
We focus on CLIP in the main text, while additional details on SAE configurations and evaluation, along with results on other ViT-based vision encoders, are provided in Appendix.

\noindent 
Based on the outlier token definition, we set the norm threshold $\tau$ to 60 for CLIP-B/16 and 100 for other models. For all SCC experiments, we set the shifting factor $s=1$, and used $k_{CDS}=5$ images for CDS calculation.

\subsection{Spatial Instability of Outliers}
\label{sec:pcp_experiment}
To empirically quantify the instability of outlier tokens, we apply the analysis described in Section \ref{sec:PCP}. 
We conduct the analysis on 1,000 images, sampled one per class from the ImageNet-1K validation set. 
In addition to measuring the average instability across all heads, we also examine per-head EMD scores for known semantic heads \cite{madasu2025pruning}, which attend strongly to meaningful objects.

\noindent
The results are summarized in Table \ref{tab:emd_instability}. 
They show that outlier tokens are highly unstable under contextual shifts. 
This gap becomes even more pronounced when the analysis is restricted to the semantic head. 
These findings suggest that outlier status is not an intrinsic property of a patch, but depends strongly on its surrounding context. 
Consequently, even subtle changes in neighboring patches can trigger a reassignment of outlier status and substantially alter the model's attention patterns.

\subsection{Feature Partitioning by CDS}
\label{sec:partition}

\begin{table}[t]
\centering
\caption{\textbf{Quantitative analysis of attention instability for outlier vs. non-outlier tokens.} We report the EMD scores ($\bar{D}_{\text{out}}$ vs. $\bar{D}_{\text{non}}$) computed on the final layer's attention maps using our SCC method with a minimal shift $s=1$ to induce perturbation. The comparison is provided for three CLIP models, averaging across all heads and isolating semantic heads.}
\label{tab:emd_instability}
\vspace{-0.2cm}
\resizebox{\columnwidth}{!}{%
\begin{tabular}{llcc}
\hline
\multirow{2}{*}{Model} & \multirow{2}{*}{Head} & \multicolumn{2}{c}{EMD} \\ \cline{3-4} 
                       &                       & non-outlier        & outlier           \\ \hline
\multirow{2}{*}{CLIP-B/16} & H11 ($D_{11}$)       & $1.365_{\pm .28}$  & $5.415_{\pm 1.52}$ \\
                           & All heads ($\bar{D}$) & $2.495_{\pm .59}$  & $5.389_{\pm 1.54}$ \\ \hline
\multirow{2}{*}{CLIP-L/14} & H4 ($D_{4}$)         & $0.796_{\pm .41}$  & $4.398_{\pm 3.33}$ \\
                           & All heads ($\bar{D}$) & $1.329_{\pm 1.20}$ & $4.411_{\pm 3.35}$ \\ \hline
\multirow{2}{*}{CLIP-L/14-336px} & H4 ($D_{4}$)   & $0.903_{\pm .47}$  & $5.164_{\pm 4.21}$ \\
                                 & All heads ($\bar{D}$) & $1.490_{\pm 1.40}$ & $5.179_{\pm 4.17}$ \\ \hline
\end{tabular}%
}
\end{table}

% \begin{table}[]
% \caption{\textbf{Feature set activations by token type.} We compare the average activation of the low-CDS set and high-CDS set in response to outlier tokens versus non-outlier tokens.}
% \resizebox{\columnwidth}{!}{%
% \begin{tabular}{llcc}
% \hline
% Model      & Feature Set                                   & Non-Outlier Token                                  & Outlier Token                                   \\ \hline
% CLIP-B/16       & \begin{tabular}[c]{@{}l@{}}high-CDS\\ low-CDS\end{tabular} &               \begin{tabular}[c]{@{}l@{}}1.66_{\pm .97}\\31.51_{\pm 4.02}\end{tabular} & \begin{tabular}[c]{@{}l@{}}83.45_{\pm 3.19}\\                 10.39_{\pm 3.22}\end{tabular}  \\ \hline
% CLIP-L/14       & \begin{tabular}[c]{@{}l@{}}high-CDS\\ low-CDS\end{tabular} &               \begin{tabular}[c]{@{}l@{}}2.69_{\pm .56}\\52.44_{\pm 5.77}\end{tabular} & \begin{tabular}[c]{@{}l@{}}171.55_{\pm 16.47}\\                25.40_{\pm 5.66}\end{tabular}  \\ \hline
% CLIP-L/14-336px & \begin{tabular}[c]{@{}l@{}}high-CDS\\ low-CDS\end{tabular} & \begin{tabular}[c]{@{}l@{}}8.94_{\pm 1.50}\\ 50.89_{\pm 5.11}\end{tabular} & \begin{tabular}[c]{@{}l@{}}173.65_{\pm 16.22}\\ 89.31_{\pm 8.64}\end{tabular} \\ \hline
% \end{tabular}%
% }
% \label{table:activation_compare}
% \vspace{-0.5cm}
% \end{table}

\begin{table}[t]
\centering
\caption{\textbf{Feature set activations by token type.} We compare the average activation of the low-CDS set and high-CDS set in response to outlier tokens versus non-outlier tokens.}
\label{table:activation_compare}
\vspace{-0.2cm}
\resizebox{\columnwidth}{!}{%
\begin{tabular}{llcc}
\hline
Model & Feature Set & Non-Outlier Token & Outlier Token \\ \hline
\multirow{2}{*}{CLIP-B/16} & high-CDS & $1.66_{\pm .97}$ & $83.45_{\pm 3.19}$ \\
 & low-CDS & $31.51_{\pm 4.02}$ & $10.39_{\pm 3.22}$ \\ \hline
\multirow{2}{*}{CLIP-L/14} & high-CDS & $2.69_{\pm .56}$ & $171.55_{\pm 16.47}$ \\
 & low-CDS & $52.44_{\pm 5.77}$ & $25.40_{\pm 5.66}$ \\ \hline
\multirow{2}{*}{CLIP-L/14-336px} & high-CDS & $8.94_{\pm 1.50}$ & $173.65_{\pm 16.22}$ \\
 & low-CDS & $50.89_{\pm 5.11}$ & $89.31_{\pm 8.64}$ \\ \hline
\end{tabular}%
}
\end{table}

% \vspace{-1cm}

\noindent
We computed the CDS for every SAE feature following the procedure described in Section \ref{sec:CIS}. 
The $k_{CDS}$ images used for this calculation were selected from the ImageNet-1K \cite{deng2009imagenet} validation set. 
The resulting CDS histograms are shown in Figure \ref{fig:cis_result}. 
Across all three CLIP models, most SAE features cluster into a dominant low-CDS peak, while distinct additional peaks appear in the high-CDS region. 
This multi-modal distribution motivates a partition of SAE features into two groups using a threshold $\gamma$ that separates the dominant low-CDS peak from the subsequent high-CDS peaks. 
We set $\gamma$ to 0.14, 0.20, and 0.13 for CLIP-B/16, CLIP-L/14, and CLIP-L/14-336px, respectively.
Features with $\text{CDS} \le \gamma$ are assigned to the low-CDS set, while those with $\text{CDS} > \gamma$ are assigned to the high-CDS set.

\noindent
We then examined how these two feature groups relate to outlier and non-outlier tokens. 
Table \ref{table:activation_compare} shows a clear contrast in activation patterns. 
For non-outlier tokens, the high-CDS set exhibits substantially lower activation than the low-CDS set. 
In contrast, for outlier tokens, the high-CDS set shows markedly stronger activation than the low-CDS set. 
This pattern indicates that the high-CDS set responds preferentially to outlier tokens, whereas the low-CDS set is more strongly associated with non-outlier tokens.

\noindent
This observation aligns with the established role of outlier tokens as global information aggregators \cite{darcet2023vision, jiang2025vision}. 
The predominant activation of the high-CDS set on outlier tokens suggests that this feature group captures broader contextual information. 
Conversely, the stronger response of the low-CDS set to non-outlier tokens suggests that it encodes more localized, patch-level information.

\subsection{Feature-Group Removal Analysis}
\label{sec:ablation}

\textbf{Removing low-CDS and high-CDS feature groups.}
To evaluate the functional roles of the low-CDS and high-CDS feature groups, we conducted feature-group removal experiments across three vision tasks: ImageNet classification \cite{deng2009imagenet}, ADE20K semantic segmentation \cite{zhou2017scene}, and NYUd monocular depth estimation \cite{Silberman:ECCV12}. 
In all experiments, we used visual tokens extracted from CLIP’s Vision Transformer as input features and trained task-specific linear probes while keeping the backbone frozen.

\begin{table}[]
\caption{\textbf{Linear probing performance of ablated feature sets.} We report linear probing results (ImageNet1K Top-1, ADE20K mIoU, NYUd RMSE) across different CLIP backbones. We compare the performance of the original embedding against two ablations: removing low-CDS features and removing high-CDS features. \textbf{BOLD} denote the best result, while \underline{UNDERLINE} denote the worst result.}
\resizebox{\columnwidth}{!}{%
\begin{tabular}{llccc}
\hline
Model               & Embedding Type                                                                    & \multicolumn{1}{c}{\begin{tabular}[c]{@{}c@{}}ImageNet1K\\ Top-1↑\end{tabular}} & \multicolumn{1}{c}{\begin{tabular}[c]{@{}c@{}}ADE20K\\ mIoU↑\end{tabular}} & \multicolumn{1}{c}{\begin{tabular}[c]{@{}c@{}}NYUd\\ rmse↓\end{tabular}} \\ \hline
CLIP-B/16       & \begin{tabular}[c]{@{}l@{}}Original\\ high-CDS-removed\\ low-CDS-removed\end{tabular} & \begin{tabular}[c]{@{}l@{}}74.82\\ \textbf{75.54}\\ \underline{64.86}\end{tabular}                   & \begin{tabular}[c]{@{}l@{}}25.87\\ \textbf{26.02}\\ \underline{11.65}\end{tabular}              
& \begin{tabular}[c]{@{}l@{}}0.8841\\ \textbf{0.8616}\\ \underline{0.9481}\end{tabular}                                                                         \\ \hline
CLIP-L/14       & \begin{tabular}[c]{@{}l@{}}Original\\ high-CDS-removed\\ low-CDS-removed\end{tabular} & \begin{tabular}[c]{@{}l@{}}80.82\\ \textbf{81.28}\\ \underline{78.30}\end{tabular}                   & \begin{tabular}[c]{@{}l@{}}\textbf{26.66}\\ 26.44\\ \underline{13.89}\end{tabular}              
& \begin{tabular}[c]{@{}l@{}}0.8029\\ \textbf{0.7994}\\ \underline{0.8878}\end{tabular}                                                                         \\ \hline
CLIP-L/14-336px & \begin{tabular}[c]{@{}l@{}}Original\\ high-CDS-removed\\ low-CDS-removed\end{tabular} & \begin{tabular}[c]{@{}l@{}}\underline{80.69}\\ 82.11\\ \textbf{82.89}\end{tabular}                   & \begin{tabular}[c]{@{}l@{}}\textbf{24.89}\\ 24.00\\ \underline{15.48}\end{tabular}              
& \begin{tabular}[c]{@{}l@{}}0.8063\\ \textbf{0.7894}\\ \underline{0.8825}\end{tabular}                                                                         \\ \hline
\end{tabular}%
}
\label{table:ablation1}
\vspace{-0.5cm}
\end{table}

\noindent
For each task, we first trained linear probes on the \textbf{original CLIP patch embeddings} as a baseline. 
We then constructed two additional embeddings by removing either the low-CDS or high-CDS feature set. 
The \textbf{low-CDS-removed embedding} was obtained by subtracting the embedding reconstructed from the low-CDS set from the original embedding, thereby removing local, patch-level information while preserving broader contextual information.
Conversely, the \textbf{high-CDS-removed embedding} was obtained by subtracting the embedding reconstructed from the high-CDS set, thereby removing broader contextual information while retaining local details.

% \begin{figure*}[htbp]
%     \centering
%     % 0.8\linewidth 또는 0.9\linewidth 등 원하는 비율로 조절
%     % 1.0\linewidth로 꽉 채우면 답답해 보일 수 있으니 0.8~0.9가 적절
%     % \includegraphics[width=1\linewidth]{image/recon_norm2.png} 
%     \includegraphics[width=1\linewidth]{image/visualization_CR.png} 
    
%     % \caption은 figure*에 직접 답니다.
%     \caption{\textbf{Visualization of per-patch $L_2$ norms for the original and feature-removed embeddings.} We visualize the per-patch $L_2$ norms of the original embedding, the low-CDS-removed embedding, and the high-CDS-removed embedding. The comparison highlights how each feature group affects outlier-token norms and the overall spatial distribution of patch norms across the three CLIP models.}
%     \label{fig:norm_vis} % 레이블도 의미있게 변경
%     \vspace{-0.5cm}
% \end{figure*}

% \noindent
% (1) Low-CDS-removed embedding: we subtracted the embedding reconstructed from the low-CDS set from the original embedding. 
% This operation effectively removes local, patch-level information while preserving broader contextual information.

% \noindent
% (2) High-CDS-removed embedding: we subtracted the embedding reconstructed from the high-CDS set from the original embedding, thereby removing broader contextual information while retaining local details.

\noindent
To ensure a fair comparison, all three embeddings were normalized before being passed to the linear probe. 
We trained linear probes on all downstream tasks using these three input embeddings. 
% Detailed methodologies and training protocols for each task are provided in Appendix.
Detailed methodologies and training protocols for each task are provided in Appendix \ref{sec:linear_probe_setup}.

\noindent
%% 여기 부분은 좀 손을 봐야할듯...
\textbf{Analysis of Linear Probing on Image Classification.}
Table \ref{table:ablation1} summarizes the linear-probe results across all CLIP backbones.
For ImageNet classification, the high-CDS-removed embedding consistently outperforms the original embedding across all three models.
In contrast, the low-CDS-removed embedding degrades performance for CLIP-B/16 and CLIP-L/14, but improves performance for CLIP-L/14-336px.
These results suggest that the two CDS-defined feature groups play different roles in linear probing.
Reducing the contribution of the high-CDS set does not hurt classification accuracy in any of the evaluated CLIP models, and in fact improves it consistently.
One possible explanation is that suppressing high-CDS features yields a representation that is easier for a linear classifier to exploit, even if those features may still contain useful information.

\begin{figure}[htbp]
    \centering
    % 0.8\linewidth 또는 0.9\linewidth 등 원하는 비율로 조절
    % 1.0\linewidth로 꽉 채우면 답답해 보일 수 있으니 0.8~0.9가 적절
    % \includegraphics[width=1\linewidth]{image/recon_norm2.png} 
    \includegraphics[width=1\linewidth]{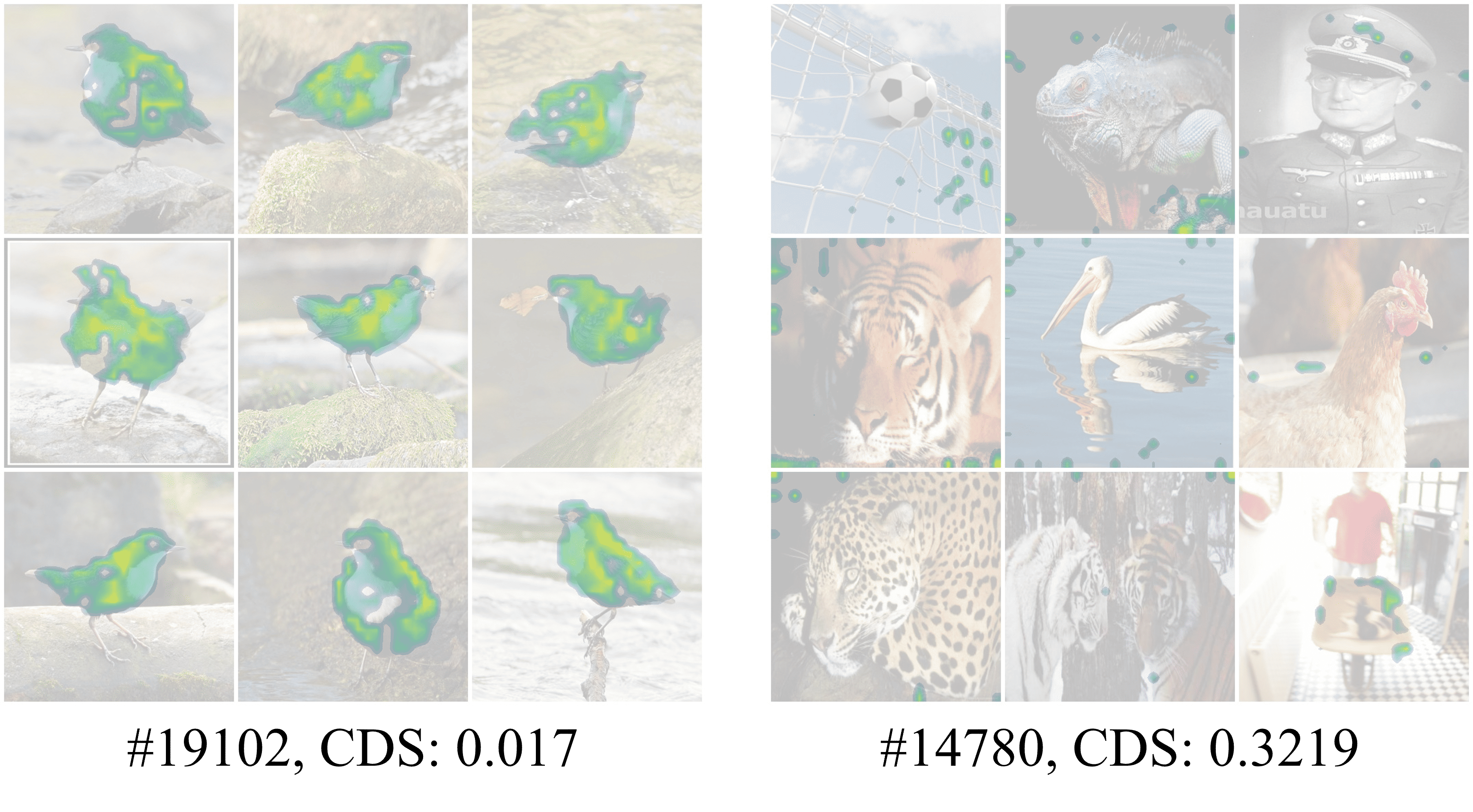} 
    
    % \caption은 figure*에 직접 답니다.
    \vspace{-0.3cm}
    \caption{\textbf{Representative activation patterns of low-CDS and high-CDS SAE features.} Low-CDS features (left) show spatially grounded and semantically consistent responses, whereas high-CDS features (right) exhibit more diffuse activations and weaker semantic consistency.}
    \label{fig:CDS_compare} % 레이블도 의미있게 변경
    \vspace{-0.5cm}
\end{figure}

\noindent
By contrast, reducing the contribution of the low-CDS set is harmful for the smaller models, indicating that low-CDS features provide information that is useful for classification.

\noindent
The CLIP-L/14-336px results suggest a more nuanced picture.
For this model, the low-CDS-removed embedding performs better than the original embedding, implying that its remaining high-CDS-dominated representation is highly discriminative for linear classification.
% This interpretation is further supported by the additional results in Appendix, where the performance gap between the original and low-CDS-removed embeddings becomes smaller as model size increases.
This interpretation is further supported by the scaling results in Appendix \ref{sec:add_result} (Figure \ref{fig:classification}), where the performance gap between the original and low-CDS-removed embeddings generally narrows with model scale.
Taken together, these results suggest that the classification utility of high-CDS features may increase with model scale or input resolution, rather than being uniformly negligible.
At the same time, the fact that high-CDS removal still improves performance indicates that, in the linear-probe setting, these features are not always the easiest for a linear classifier to exploit directly when mixed with low-CDS features in the original representation.

\begin{figure*}[htbp]
    \centering
    % 0.8\linewidth 또는 0.9\linewidth 등 원하는 비율로 조절
    % 1.0\linewidth로 꽉 채우면 답답해 보일 수 있으니 0.8~0.9가 적절
    % \includegraphics[width=1\linewidth]{image/recon_norm2.png} 
    \includegraphics[width=1\linewidth]{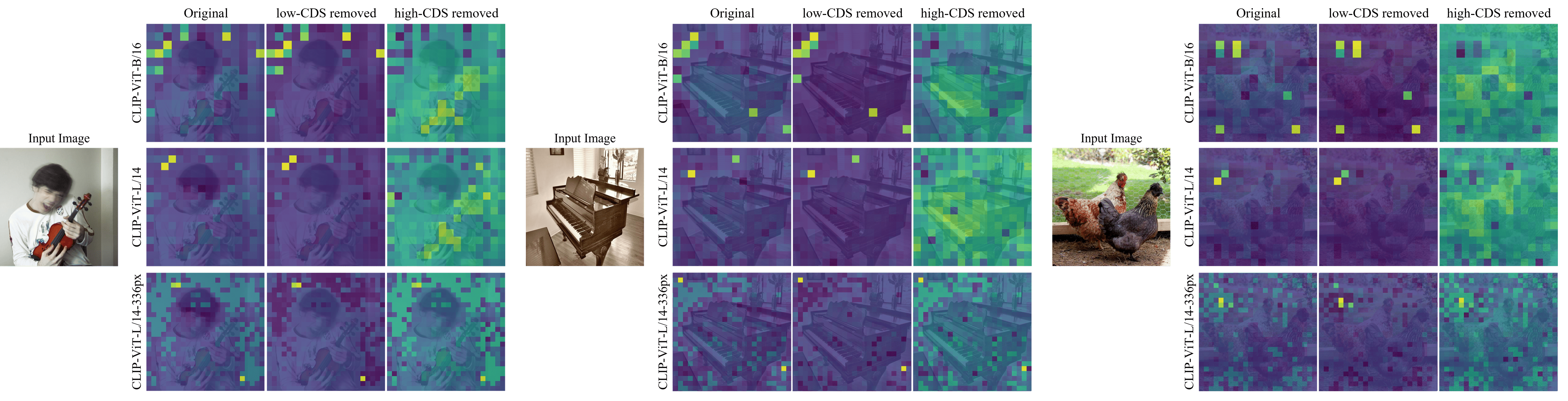} 
    
    % \caption은 figure*에 직접 답니다.
    \caption{\textbf{Visualization of per-patch $L_2$ norms for the original and feature-removed embeddings.} We visualize the per-patch $L_2$ norms of the original embedding, the low-CDS-removed embedding, and the high-CDS-removed embedding. The comparison highlights how each feature group affects outlier-token norms and the overall spatial distribution of patch norms across the three CLIP models.}
    \label{fig:norm_vis} % 레이블도 의미있게 변경
    \vspace{-0.5cm}
\end{figure*}

\noindent
\textbf{Analysis of Linear Probing on Dense Prediction Tasks.}
We next evaluate the representations on dense prediction tasks, namely ADE20K semantic segmentation and NYUd monocular depth estimation.
Across all experiments, reducing the contribution of the low-CDS set causes substantial performance degradation.
These results provide strong evidence that low-CDS feature capture local information essential for dense prediction.

\noindent
The high-CDS-removed condition provides a perspective on the role of broader contextual information across the two tasks. 
On ADE20K, where the linear probe operates only on patch representations, the high-CDS-removed embedding performs comparably to or slightly better than the original baseline. 
This suggests that the information encoded by the high-CDS set contributes only marginally to pixel-level semantic prediction, which is governed primarily by local cues.
In contrast, on NYUd, the high-CDS-removed embedding also improves performance over the original baseline.
One possible explanation is that scene-level contextual information remains available after high-CDS removal because the CLS token is concatenated to every patch token prior to upsampling.
This may allow the linear probe to retain global cues while operating on a cleaner patch-focused representation.
Taken together, these results indicate that low-CDS features provide the local information required for dense prediction, whereas the high-CDS set mainly contributes complementary broader context.
% Similar trends are also observed on other ViTs in Appendix \ref{sec:add_result} (Table \ref{table:ablation_vit}), where low-CDS removal consistently causes large degradation, while high-CDS removal remains close to the original embedding.

\subsection{Qualitative Feature Analysis}

% \begin{figure}[htbp]
%     \centering
%     % 0.8\linewidth 또는 0.9\linewidth 등 원하는 비율로 조절
%     % 1.0\linewidth로 꽉 채우면 답답해 보일 수 있으니 0.8~0.9가 적절
%     % \includegraphics[width=1\linewidth]{image/recon_norm2.png} 
%     \includegraphics[width=1\linewidth]{image/CDS_compare.png} 
    
%     % \caption은 figure*에 직접 답니다.
%     \caption{\textbf{Representative activation patterns of low-CDS and high-CDS SAE features.}}
%     \label{fig:CDS_compare} % 레이블도 의미있게 변경
%     \vspace{-0.5cm}
% \end{figure}

\textbf{Visualizing Low- and High-CDS Features.}
To further examine what CDS captures, we visualize representative SAE features learned from CLIP-L/14-336px. 
Figure \ref{fig:CDS_compare} compares representative low-CDS and high-CDS features in terms of spatial grounding and semantic consistency. 
Low-CDS features exhibit strong spatial grounding, with activation maps that are highly localized to specific objects or visual attributes within the image. 
Their top-activating images also show high semantic consistency, suggesting that low-CDS features are closely tied to distinct visual concepts.

In contrast, high-CDS features display more diffuse and ambiguous activation patterns, rarely localizing to a specific region. 
Compared with low-CDS features, their top-activating images also appear less semantically consistent in the representative examples shown in Figure \ref{fig:CDS_compare}. 
Additional examples in Appendix (Figures \ref{fig:low_cds_visual} and \ref{fig:high_cds_visual}) further suggest that this tendency is not uniform. 
Some high-CDS features remain broadly related across images, whereas others activate on image sets with weaker semantic coherence.

Overall, these visualizations support our interpretation of CDS as a measure of information scope. 
Low-CDS features behave like localized detectors, whereas high-CDS features reflect broader and more context-dependent representations.

\noindent
\textbf{Visualization of Feature Removal.}
Figure~\ref{fig:norm_vis} visualizes the per-patch $L_2$ norm of the original embedding, the low-CDS-removed embedding, and the high-CDS-removed embedding for an input image. 
Across all models, the low-CDS-removed embeddings largely retain the high norm values at outlier-token locations compared with the original embeddings. 
This suggests that the features primarily responsible for forming outlier tokens are largely drawn from the high-CDS set, consistent with the results in Table \ref{table:activation_compare}.

\noindent
For CLIP-B/16 and CLIP-L/14, the high-CDS-removed embeddings show a clear suppression of norm values at the original outlier-token locations. 
Instead, these embeddings exhibit relatively elevated norms across the remaining patches, especially around semantic objects.
By contrast, the high-CDS-removed embedding of CLIP-L/14-336px still retains relatively high norms at the outlier-token locations. 
This qualitative pattern is consistent with Table \ref{table:activation_compare}, which shows comparatively strong low-CDS feature activation within outlier tokens for this model.

\noindent
To better understand this exception, we examined the CLIP-L/14-336px visualizations more closely. 
Unlike the other models, CLIP-L/14-336px exhibits elevated norm values not only at outlier-token locations but also across many background patches. 
This elevated background norm remains particularly visible in the high-CDS-removed embedding, suggesting that the model has learned localized low-CDS features associated with background regions.
Because outlier tokens often emerge in visually homogeneous background areas, this observation suggests that the outlier tokens of CLIP-L/14-336px share a substantial portion of these background-related low-CDS features.

\section{Conclusion}
We introduced information scope as a new axis for interpreting CLIP representations through SAE features.
To quantify it, we proposed the Contextual Dependency Score (CDS) and applied it to multiple CLIP vision encoders.
Our results reveal a consistent separation between low-CDS and high-CDS features and show that they play distinct roles across downstream tasks.
Overall, information scope provides a useful complementary perspective beyond semantic identity for analyzing SAE features in CLIP.
% Additional results in the Appendix further suggest that this perspective may generalize beyond CLIP to other ViT-based vision encoders.
\section{Acknowledgments}

This work was partly supported by the Institute of Information \& Communications Technology Planning \& Evaluation (IITP) grant funded by the Korea government (MSIT) (RS-2025-02283048, Developing the Next-Generation General AI with Reliability, Ethics, and Adaptability, 70\%) and partly supported by Basic Science Research Program through the 
National Research Foundation of Korea (NRF) funded by the Ministry of Education (RS-2025-25418053, 30\%).
{
    \small
    \bibliographystyle{ieeenat_fullname}
    \bibliography{main}

@String(ECCV= {Eur. Conf. Comput. Vis.})

@String(BMVC= {Brit. Mach. Vis. Conf.})

@String(ECCV  = {ECCV})

@String(BMVC  =	{BMVC})

@article{darcet2023vision,
  title={Vision transformers need registers},
  author={Darcet, Timoth{\'e}e and Oquab, Maxime and Mairal, Julien and Bojanowski, Piotr},
  journal={arXiv preprint arXiv:2309.16588},
  year={2023}
}

@article{jiang2025vision,
  title={Vision Transformers Don't Need Trained Registers},
  author={Jiang, Nick and Dravid, Amil and Efros, Alexei and Gandelsman, Yossi},
  journal={arXiv preprint arXiv:2506.08010},
  year={2025}
}

@article{pach2025sparse,
  title={Sparse autoencoders learn monosemantic features in vision-language models},
  author={Pach, Mateusz and Karthik, Shyamgopal and Bouniot, Quentin and Belongie, Serge and Akata, Zeynep},
  journal={arXiv preprint arXiv:2504.02821},
  year={2025}
}

@article{lim2024sparse,
  title={Sparse autoencoders reveal selective remapping of visual concepts during adaptation},
  author={Lim, Hyesu and Choi, Jinho and Choo, Jaegul and Schneider, Steffen},
  journal={arXiv preprint arXiv:2412.05276},
  year={2024}
}

@inproceedings{radford2021learning,
  title={Learning transferable visual models from natural language supervision},
  author={Radford, Alec and Kim, Jong Wook and Hallacy, Chris and Ramesh, Aditya and Goh, Gabriel and Agarwal, Sandhini and Sastry, Girish and Askell, Amanda and Mishkin, Pamela and Clark, Jack and others},
  booktitle={International conference on machine learning},
  pages={8748--8763},
  year={2021},
  organization={PmLR}
}

@article{bussmann2024batchtopk,
  title={Batchtopk sparse autoencoders},
  author={Bussmann, Bart and Leask, Patrick and Nanda, Neel},
  journal={arXiv preprint arXiv:2412.06410},
  year={2024}
}

@article{han2025causal,
  title={Causal Interpretation of Sparse Autoencoder Features in Vision},
  author={Han, Sangyu and Kim, Yearim and Kwak, Nojun},
  journal={arXiv preprint arXiv:2509.00749},
  year={2025}
}

@article{zaigrajew2025interpreting,
  title={Interpreting CLIP with Hierarchical Sparse Autoencoders},
  author={Zaigrajew, Vladimir and Baniecki, Hubert and Biecek, Przemyslaw},
  journal={arXiv preprint arXiv:2502.20578},
  year={2025}
}

@inproceedings{rao2024discover,
  title={Discover-then-name: Task-agnostic concept bottlenecks via automated concept discovery},
  author={Rao, Sukrut and Mahajan, Sweta and B{\"o}hle, Moritz and Schiele, Bernt},
  booktitle={European Conference on Computer Vision},
  pages={444--461},
  year={2024},
  organization={Springer}
}

@article{gao2024scaling,
  title={Scaling and evaluating sparse autoencoders},
  author={Gao, Leo and la Tour, Tom Dupr{\'e} and Tillman, Henk and Goh, Gabriel and Troll, Rajan and Radford, Alec and Sutskever, Ilya and Leike, Jan and Wu, Jeffrey},
  journal={arXiv preprint arXiv:2406.04093},
  year={2024}
}

@article{cunningham2023sparse,
  title={Sparse autoencoders find highly interpretable features in language models},
  author={Cunningham, Hoagy and Ewart, Aidan and Riggs, Logan and Huben, Robert and Sharkey, Lee},
  journal={arXiv preprint arXiv:2309.08600},
  year={2023}
}

@article{olah2017feature,
  author = {Olah, Chris and Mordvintsev, Alexander and Schubert, Ludwig},
  title = {Feature Visualization},
  journal = {Distill},
  year = {2017},
  note = {https://distill.pub/2017/feature-visualization},
  doi = {10.23915/distill.00007}
}

@article{dosovitskiy2020image,
  title={An image is worth 16x16 words: Transformers for image recognition at scale},
  author={Dosovitskiy, Alexey},
  journal={arXiv preprint arXiv:2010.11929},
  year={2020}
}

@article{olshausen1997sparse,
  title={Sparse coding with an overcomplete basis set: A strategy employed by V1?},
  author={Olshausen, Bruno A and Field, David J},
  journal={Vision research},
  volume={37},
  number={23},
  pages={3311--3325},
  year={1997},
  publisher={Elsevier}
}

@article{fry2024towards,
  title={Towards multimodal interpretability: Learning sparse interpretable features in vision transformers, 2024},
  author={Fry, Hugo},
  journal={URL https://www. lesswrong. com/posts/bCtbuWraqYTDtuARg/towards-multimodal-interpretability-learning-sparse-2\# Future\_ Work. Accessed},
  pages={06--28},
  year={2024}
}

@article{daujotas2024interpreting,
  title={Interpreting and steering features in images},
  author={Daujotas, Gytis},
  journal={URL https://www. lesswrong. com/posts/Quqekpvx8BGMMcaem/interpreting-andsteering-features-in-images},
  volume={4},
  year={2024}
}

@article{daujotas2024case,
  title={Case study: Interpreting, manipulating, and controlling clip with sparse autoencoders, 2024},
  author={Daujotas, Gytis},
  journal={URL https://www. lesswrong. com/posts/iYFuZo9BMvr6GgMs5/case-study-interpreting-manipulating-and-controlling-clip. Accessed},
  pages={09--24},
  year={2024}
}

@inproceedings{thasarathan2025universal,
  title={Universal sparse autoencoders: Interpretable cross-model concept alignment},
  author={Thasarathan, Harrish and Forsyth, Julian and Fel, Thomas and Kowal, Matthew and Derpanis, Konstantinos G},
  booktitle={Forty-second International Conference on Machine Learning},
  year={2025}
}

@article{oquab2023dinov2,
  title={Dinov2: Learning robust visual features without supervision},
  author={Oquab, Maxime and Darcet, Timoth{\'e}e and Moutakanni, Th{\'e}o and Vo, Huy and Szafraniec, Marc and Khalidov, Vasil and Fernandez, Pierre and Haziza, Daniel and Massa, Francisco and El-Nouby, Alaaeldin and others},
  journal={arXiv preprint arXiv:2304.07193},
  year={2023}
}

@article{rubner2000earth,
  title={The earth mover's distance as a metric for image retrieval},
  author={Rubner, Yossi and Tomasi, Carlo and Guibas, Leonidas J},
  journal={International journal of computer vision},
  volume={40},
  number={2},
  pages={99--121},
  year={2000},
  publisher={Springer}
}

@inproceedings{deng2009imagenet,
  title={Imagenet: A large-scale hierarchical image database},
  author={Deng, Jia and Dong, Wei and Socher, Richard and Li, Li-Jia and Li, Kai and Fei-Fei, Li},
  booktitle={2009 IEEE conference on computer vision and pattern recognition},
  pages={248--255},
  year={2009},
  organization={Ieee}
}

@inproceedings{zhou2017scene,
  title={Scene parsing through ade20k dataset},
  author={Zhou, Bolei and Zhao, Hang and Puig, Xavier and Fidler, Sanja and Barriuso, Adela and Torralba, Antonio},
  booktitle={Proceedings of the IEEE conference on computer vision and pattern recognition},
  pages={633--641},
  year={2017}
}

@inproceedings{Silberman:ECCV12,
  author    = {Nathan Silberman, Derek Hoiem, Pushmeet Kohli and Rob Fergus},
  title     = {Indoor Segmentation and Support Inference from RGBD Images},
  booktitle = {ECCV},
  year      = {2012}
}

@article{madasu2025pruning,
  title={Pruning the Paradox: How CLIP's Most Informative Heads Enhance Performance While Amplifying Bias},
  author={Madasu, Avinash and Lal, Vasudev and Howard, Phillip},
  journal={arXiv preprint arXiv:2503.11103},
  year={2025}
}

@inproceedings{bhat2021adabins,
  title={Adabins: Depth estimation using adaptive bins},
  author={Bhat, Shariq Farooq and Alhashim, Ibraheem and Wonka, Peter},
  booktitle={Proceedings of the IEEE/CVF conference on computer vision and pattern recognition},
  pages={4009--4018},
  year={2021}
}

@article{eigen2014depth,
  title={Depth map prediction from a single image using a multi-scale deep network},
  author={Eigen, David and Puhrsch, Christian and Fergus, Rob},
  journal={Advances in neural information processing systems},
  volume={27},
  year={2014}
}

@article{bricken2023monosemanticity,
   title={Towards Monosemanticity: Decomposing Language Models With Dictionary Learning},
   author={Bricken, Trenton and Templeton, Adly and Batson, Joshua and Chen, Brian and Jermyn, Adam and Conerly, Tom and Turner, Nick and Anil, Cem and Denison, Carson and Askell, Amanda and Lasenby, Robert and Wu, Yifan and Kravec, Shauna and Schiefer, Nicholas and Maxwell, Tim and Joseph, Nicholas and Hatfield-Dodds, Zac and Tamkin, Alex and Nguyen, Karina and McLean, Brayden and Burke, Josiah E and Hume, Tristan and Carter, Shan and Henighan, Tom and Olah, Christopher},
   year={2023},
   journal={Transformer Circuits Thread},
   note={https://transformer-circuits.pub/2023/monosemantic-features/index.html}
}

@article{park2025decoding,
  title={Decoding Dense Embeddings: Sparse Autoencoders for Interpreting and Discretizing Dense Retrieval},
  author={Park, Seongwan and Kim, Taeklim and Ko, Youngjoong},
  journal={arXiv preprint arXiv:2506.00041},
  year={2025}
}

@article{stevens2025sparse,
  title={Sparse autoencoders for scientifically rigorous interpretation of vision models},
  author={Stevens, Samuel and Chao, Wei-Lun and Berger-Wolf, Tanya and Su, Yu},
  journal={arXiv preprint arXiv:2502.06755},
  year={2025}
}

@article{hindupur2025projecting,
  title={Projecting assumptions: The duality between sparse autoencoders and concept geometry},
  author={Hindupur, Sai Sumedh R and Lubana, Ekdeep Singh and Fel, Thomas and Ba, Demba},
  journal={arXiv preprint arXiv:2503.01822},
  year={2025}
}

@article{elhage2022toy,
  title={Toy models of superposition},
  author={Elhage, Nelson and Hume, Tristan and Olsson, Catherine and Schiefer, Nicholas and Henighan, Tom and Kravec, Shauna and Hatfield-Dodds, Zac and Lasenby, Robert and Drain, Dawn and Chen, Carol and others},
  journal={arXiv preprint arXiv:2209.10652},
  year={2022}
}

@article{bachregisters,
  title={Registers in Small Vision Transformers: A Reproducibility Study of Vision Transformers Need Registers},
  author={Bach, Linus Ruben and Bakker, Emma and van Dijk, R{\'e}nan and de Vries, Jip and Szewczyk, Konrad},
  journal={Transactions on Machine Learning Research}
}

@inproceedings{baccouche2012spatio,
  title={Spatio-Temporal Convolutional Sparse Auto-Encoder for Sequence Classification.},
  author={Baccouche, Moez and Mamalet, Franck and Wolf, Christian and Garcia, Christophe and Baskurt, Atilla},
  booktitle={BMVC},
  pages={1--12},
  year={2012}
}

@inproceedings{zhang2025large,
  title={Large multi-modal models can interpret features in large multi-modal models},
  author={Zhang, Kaichen and Shen, Yifei and Li, Bo and Liu, Ziwei},
  booktitle={Proceedings of the IEEE/CVF International Conference on Computer Vision},
  pages={3650--3661},
  year={2025}
}

@inproceedings{olson2025probing,
  title={Probing the representational power of sparse autoencoders in vision models},
  author={Olson, Matthew Lyle and Hinck, Musashi and Ratzlaff, Neale and Li, Changbai and Howard, Phillip and Lal, Vasudev and Tseng, Shao-Yen},
  booktitle={Proceedings of the IEEE/CVF International Conference on Computer Vision},
  pages={6167--6177},
  year={2025}
}

@article{tschannen2025siglip,
  title={SigLIP 2: Multilingual Vision-Language Encoders with Improved Semantic Understanding},
  author={Tschannen, Michael and Gritsenko, Alexey and Wang, Xiao and Naeem, Muhammad Ferjad and Alabdulmohsin, Ibrahim and Parthasarathy, Nikhil and Evans, Talfan and Beyer, Lucas and Xia, Ye and Mustafa, Basil and others},
  journal={Localization, and Dense Features},
  volume={6},
  year={2025}
}

@inproceedings{touvron2022deit,
  title={Deit iii: Revenge of the vit},
  author={Touvron, Hugo and Cord, Matthieu and J{\'e}gou, Herv{\'e}},
  booktitle={European conference on computer vision},
  pages={516--533},
  year={2022},
  organization={Springer}
}
}

\clearpage
\setcounter{page}{1}
\maketitlesupplementary
\appendix

\section{SAE Configuration Across Vision Encoders}
\label{sec:sae_config}
\subsection{Details for CLIP SAE Configurations}

This section provides additional statistics for the CLIP SAE configurations used in the main paper. 
We follow the same BatchTopK SAE setup for CLIP-ViT-B/16, CLIP-ViT-L/14, and CLIP-ViT-L/14-336px, and report here the resulting dictionary sizes and CDS-based partition statistics.

The resulting SAE dictionaries contain 24,576 features for CLIP-ViT-B/16 and 32,768 features for both CLIP-ViT-L/14 and CLIP-ViT-L/14-336px. 
To separate low-CDS and high-CDS features, we use the same threshold $\gamma$ as defined in the main paper. 
The resulting partition sizes are summarized in Table~\ref{tab:clip_sae_config}. 
Across all three CLIP backbones, the high-CDS set forms a relatively small subset of the overall SAE dictionary.

\begin{table}[H]
\centering
\small
\resizebox{\columnwidth}{!}{
\begin{tabular}{lcccc}
\toprule
Model & Dictionary Size & $\gamma$ & Low-CDS & High-CDS \\
\midrule
CLIP-B/16       & 24,576 & 0.14 & 17,329 & 7,247 \\
CLIP-L/14       & 32,768 & 0.20 & 27,887 & 4,881 \\
CLIP-L/14-336px & 32,768 & 0.13 & 22,323 & 10,445 \\
\bottomrule
\end{tabular}}
\caption{SAE dictionary sizes and CDS-based partition statistics for the CLIP vision encoders used in the main paper. The high-CDS set forms a relatively small subset of the full SAE dictionary across all three CLIP backbones.}
\label{tab:clip_sae_config}
\end{table}

\subsection{SAE Configuration for Other ViTs}
\label{appendix:sae_other_vit}
To assess whether our findings generalize beyond CLIP, we additionally train BatchTopK SAEs on three representative ViT families: DINOv2 \cite{oquab2023dinov2}, SigLIP2 \cite{tschannen2025siglip}, and DeiT3 \cite{touvron2022deit}. 
These experiments are intended to test whether the main phenomena studied in this paper, including outlier instability under SCC and the CDS-based separation of feature groups, also arise in other vision transformer architectures.

For each architecture, we train SAEs at three model scales for the scaling analysis in linear-probe classification. 
For all other mechanistic analyses, including the SCC-based analysis of outlier instability, we use the Base-scale models to enable a controlled comparison across architectures.

All SAEs are trained on the ImageNet-1K training set. 
During SAE training, we randomly sample four visual patch tokens per image from a designated transformer layer of each ViT, and all subsequent analyses are performed on that layer. 
For all models, we use an expansion factor of $\epsilon = 32$ and set $K = 30$ in BatchTopK.

We determine the outlier threshold $\tau$ for each model empirically from the distribution of patch-token norms. 
Table~\ref{tab:other_vit_config} summarizes the transformer layer used for SAE training and analysis, together with the corresponding outlier threshold for each model.

\begin{table}[t]
\centering
\small
\begin{tabular}{llcc}
\toprule
Architecture & Model & Layer Used & $\tau$ \\
\midrule
 & Base  & Penultimate (11th) & 100 \\
DINOv2 & Large & Penultimate (23rd) & 100 \\
 & Giant & Last (40th)        & 150 \\
\midrule
 & Small & 10th               & 500 \\
DeiT3  & Base  & 9th                & 500 \\
 & Large & 18th               & 1000 \\
\midrule
 & Base  & Last (12th)       & 150 \\
SigLIP2 & Large & Last (24th)       & 150 \\
 & Giant & Last (27th)       & 300 \\
\bottomrule
\end{tabular}
\caption{Model-specific SAE configurations for the additional ViT architectures. We report the transformer layer used for SAE training and analysis, together with the empirically selected outlier threshold $\tau$ for each model.}
\label{tab:other_vit_config}
\end{table}

\section{SAE Evaluation Across Vision Encoders}
\label{sec:sae_eval}

We have included a comprehensive evaluation of our trained SAEs in the Table \ref{appendix:sae_eval}, measuring Fraction of Variance Unexplained (FVU) as a scale-invariant measure of reconstruction error, along with $L_0$ sparsity ($L_0$), and Cosine Similarity (CS).
These metrics were computed on the ImageNet-1K validation set, covering the entire token set and the outlier subset to directly address concerns regarding high-norm tokens.

\begin{table}[H]
    \centering
    \setlength{\tabcolsep}{4pt}
    \renewcommand{\arraystretch}{1.0}
    \resizebox{\columnwidth}{!}{
    \begin{tabular}{lcccccc}
        \toprule
        Model & FVU$_A$ $\downarrow$ & FVU$_O$ $\downarrow$ & $L_0{}_A$ & $L_0{}_O$ & CS$_A$ $\uparrow$ & CS$_O$ $\uparrow$ \\
        \midrule
        CLIP-B/16        & .099 & .021 & 30.201 & 13.064 & .923 & .998 \\
        CLIP-L/14        & .176 & .042 & 29.307 & 14.170 & .873 & .999 \\
        CLIP-L/14-336px  & .115 & .028 & 29.654 & 16.391 & .913 & .999 \\
        DINOv2-B         & .039 & .001 & 29.425 &  7.298 & .911 & .999 \\
        DINOv2-L         & .115 & .001 & 30.416 & 15.334 & .917 & .999 \\
        DINOv2-G         & .062 & .002 & 30.664 & 28.709 & .912 & .999 \\
        DeiT3-S          & .123 & .081 & 32.067 & 20.318 & .941 & .956 \\
        DeiT3-B          & .215 & .101 & 32.293 & 10.208 & .900 & .940 \\
        DeiT3-L          & .170 & .096 & 31.533 & 16.113 & .930 & .953 \\
        SigLIP2-B        & .030 & .001 & 29.982 & 13.183 & .948 & .999 \\
        SigLIP2-L        & .042 & .001 & 29.628 & 13.573 & .925 & .999 \\
        SigLIP2-G        & .116 & .001 & 30.144 & 10.202 & .918 & .999 \\
        \bottomrule
    \end{tabular}}
    \caption{SAE reconstruction statistics for all tokens (A) and outlier tokens (O).}
    \label{appendix:sae_eval}
\end{table}

\section{Task-Specific Linear Probe Setup}
\label{sec:linear_probe_setup}
In this section, we provide detailed experimental setups for the downstream tasks used in Section \ref{sec:ablation} to evaluate the functional roles of the low-CDS and high-CDS feature sets.

\textbf{Probing Method.}
For ImageNet classification, we first applied global average pooling over all visual tokens after feature-group removal to obtain a single image-level embedding. A linear classifier was then trained on these pooled embeddings to predict class labels. For ADE20K semantic segmentation, we trained a linear probe on the feature-group-removed embeddings, following the protocol outlined in \cite{oquab2023dinov2}.

\noindent
For monocular depth estimation on the NYUd dataset, we first preprocessed the ground-truth (GT) depth maps. We applied a center crop to the GT maps to match the spatial resolution of the ViT input. From the final layer of the ViT, we extracted both the CLS token and the patch tokens. The corresponding feature-group removal was applied only to the patch tokens, while the CLS token remained unmodified. We then broadcast the CLS token and concatenated it to every patch token along the feature dimension. These augmented patch tokens were reshaped into their 2D spatial grid and bilinearly upsampled to the full input resolution. We formulated depth estimation as a classification task by discretizing the depth range into 256 uniformly distributed bins. A linear probe was trained on the upsampled features to predict the logits for these bins. The final scalar depth value for each pixel was computed as a linear combination of the bin centers, weighted by the predicted softmax probabilities \cite{bhat2021adabins}. The probe was trained end-to-end using a scale-invariant loss \cite{eigen2014depth} on the estimated depth.

\textbf{Training Settings.}
All linear probes were trained for 100 epochs using SGD with momentum 0.9 and a cosine annealing scheduler. The batch size was set to 1024 for ImageNet classification. For the dense prediction tasks, we used smaller batch sizes of 64 for ADE20K segmentation and 8 for NYUd depth estimation due to GPU memory constraints. We conducted a random hyperparameter search to determine the optimal learning rate and weight decay. Specifically, the learning rate was sampled from $[10^{-3}, 10^{-2}]$ for classification and segmentation, while a higher range of $[10^{-1}, 1]$ was used for depth estimation to accommodate pixel-wise optimization. The weight decay was sampled from $[10^{-5}, 10^{-4}]$ across all tasks. For each embedding variant, we performed 5 independent trials and reported the result achieving the best performance on the validation set in Table \ref{table:ablation1}.

\begin{figure*}[t]
    \centering
    \subfloat[DINOv2 \label{fig:sub1}]{
        \includegraphics[width=0.32\textwidth]{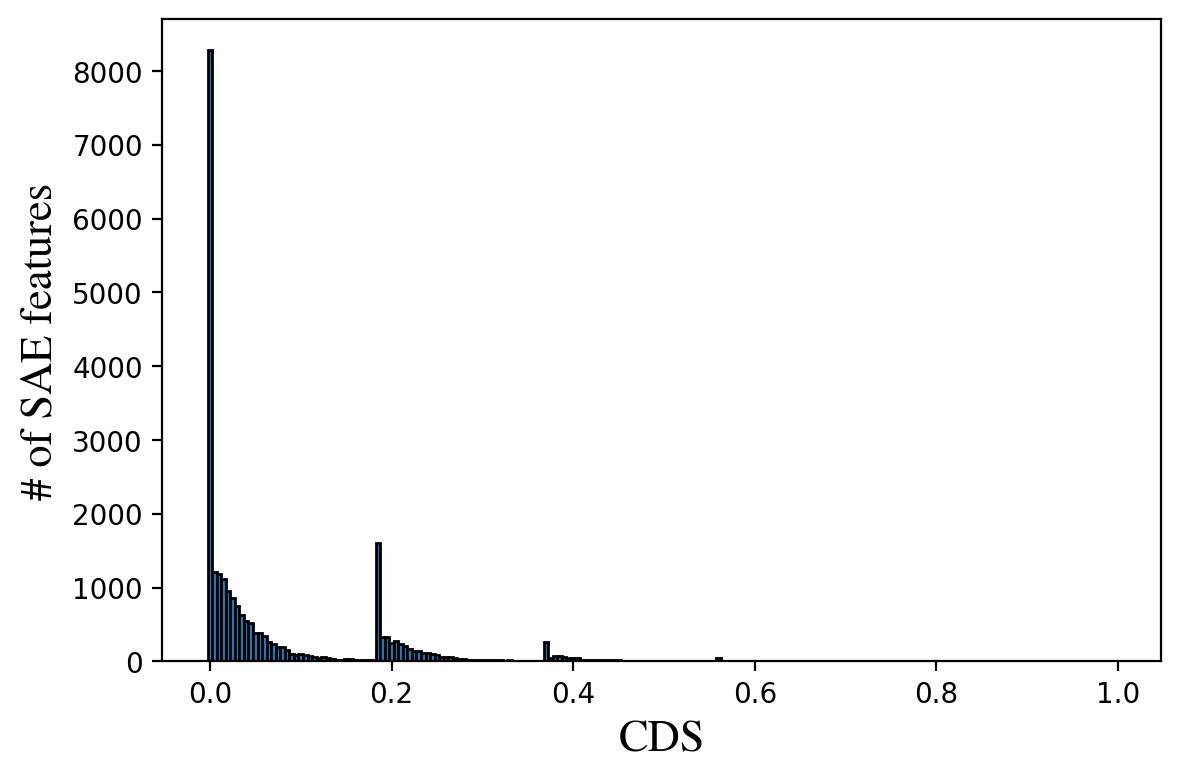}
    }
    \hfill
    \subfloat[DeiT3 \label{fig:sub2}]{
        \includegraphics[width=0.32\textwidth]{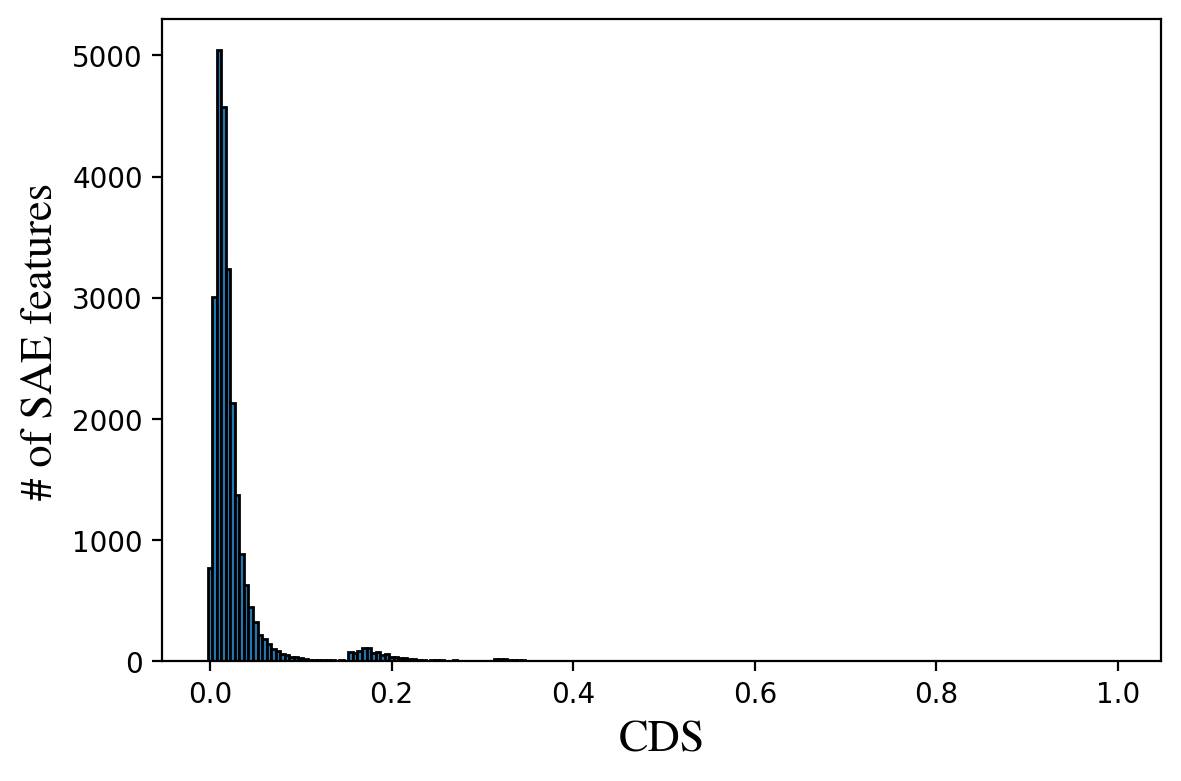}
    }
    \hfill
    \subfloat[SigLIP2 \label{fig:sub3}]{
        \includegraphics[width=0.32\textwidth]{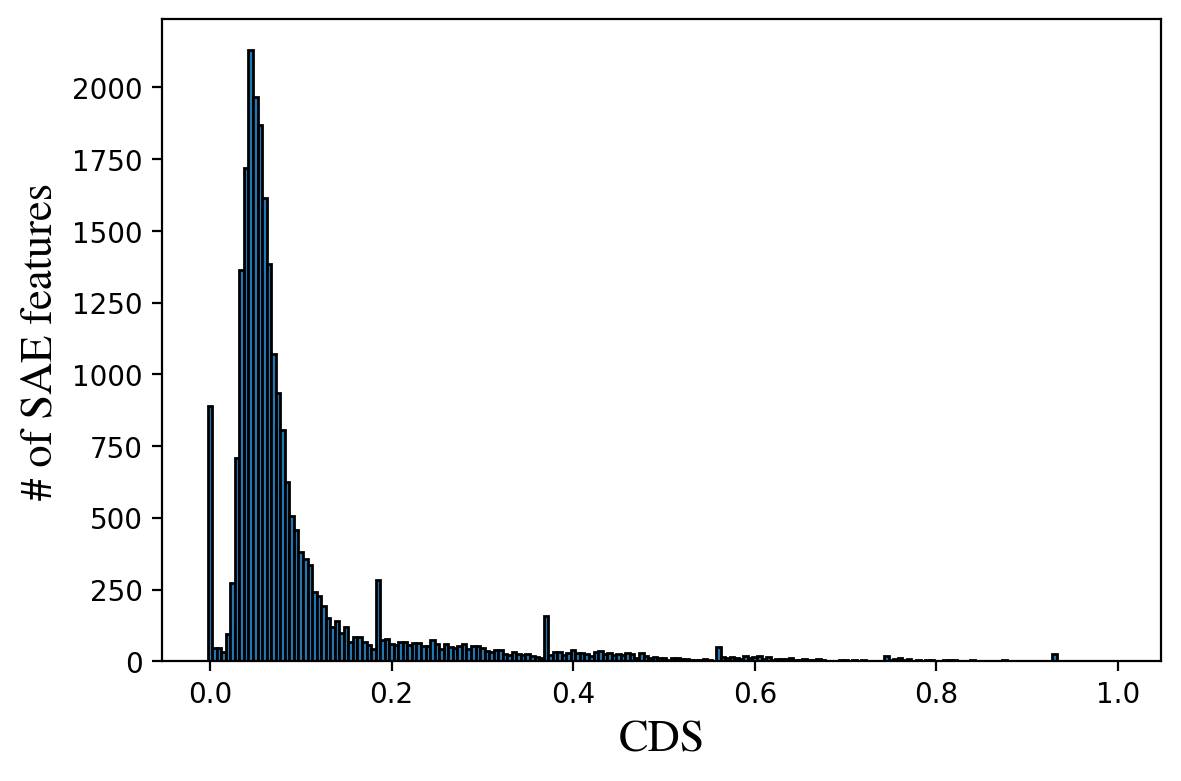}
    }
    \vspace{-0.3cm}
    \caption{\textbf{Distribution of Contextual Dependency Score (CDS).} Histogram of CDS values for SAE features across DINOv2, SigLIP2, and DeiT3.}
    \label{fig:vit_cis_result}
    \vspace{-0.5cm}
\end{figure*}

\section{Experiments on other ViTs}
\label{sec:add_result}

\subsection{CDS Distributions and Instability Across ViTs}
\label{sec:instability_analysis}
We first examine whether the CDS-based feature partition observed in CLIP also emerges in other ViT families. 
Figure \ref{fig:vit_cis_result} shows the CDS distributions of the Base-scale DINOv2, SigLIP2, and DeiT3 models. 
Across all three architectures, most SAE features cluster into a dominant low-CDS region, while additional mass appears in a separated high-CDS region. 
This qualitatively mirrors the trend observed in CLIP and supports the use of a CDS-based partition into low-CDS and high-CDS feature groups beyond CLIP.

We next ask whether outlier tokens in these architectures likewise exhibit strong contextual instability. 
To empirically quantify the spatial instability of outlier tokens, we apply the analysis detailed in Section \ref{sec:PCP}. 
We conduct our analysis using 1,000 images, randomly sampled one-per-class from the ImageNet-1K validation set. 
As reported in Table \ref{tab:emd_comparison}, outlier tokens exhibit substantially higher EMD than non-outliers across all evaluated ViTs. 
This drastic discrepancy confirms that outlier emergence is not an intrinsic property of the underlying visual patch, but rather a highly context-dependent phenomenon. 
Consequently, even a subtle spatial shift in the surrounding context can trigger a complete reassignment of outlier locations, severely disrupting the model's attention patterns.

To connect this feature-level CDS separation with token-level outlier behavior, we next investigate representational instability by quantifying contextual dependency at the token level. 

\begin{figure}[htbp]
    % --------------------------------------------------
    % 1. Table 영역 (전체 너비의 35% 할당)
    % --------------------------------------------------
    \centering
    % minipage 내부이므로 \linewidth는 할당된 60% 공간을 의미합니다.
    \includegraphics[width=\linewidth]{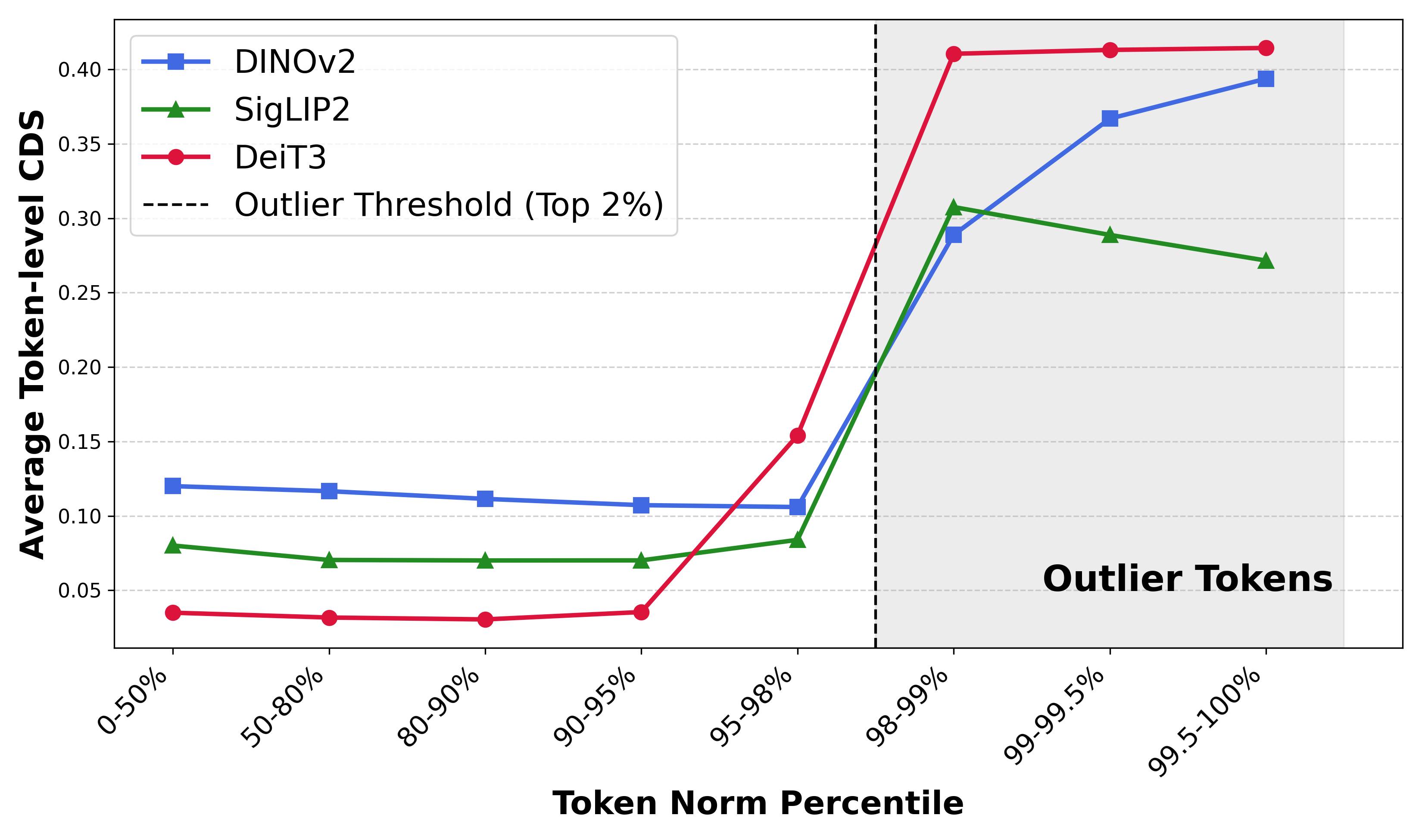} 
    % \caption{\textbf{Average Token-level CDS}}
    \caption{\textbf{Contextual dependency across the norm spectrum.} We plot the Activation-Weighted CDS ($\text{awCDS}$) across the token norm spectrum. Across all architectures, $\text{awCDS}$ spikes exponentially for outlier tokens.}
    \label{fig:Average_CDS}
\end{figure}

We define the Activation-Weighted CDS ($\mathrm{awCDS}$) of a token $v_t$ as the activation-weighted average of the CDS values of the SAE features activated by that token:
\[
\mathrm{awCDS}(v_t) = \frac{\sum_{j=1}^{q} a_{t,j}\,\mathrm{CDS}_j}{\sum_{j=1}^{q} a_{t,j}},
\]
where $a_{t,j}$ denotes the activation of the $j$-th SAE feature for token $v_t$.
% We define the Activation-Weighted CDS ($\text{awCDS}$) for a given token as the activation-weighted average of its constituent SAE features' CDS:
% $$\text{awCDS} = \frac{\sum_{j} a_j \cdot \text{CDS}_j}{\sum_{j} a_j}$$
% where $a_j$ denotes the activation value of the $j$-th SAE feature.  
To systematically analyze this metric across the token spectrum, we sort the patch tokens extracted from the same set of 1,000 images by their norm, divide them into percentile-based bins, and compute the average $\text{awCDS}$ for each bin.
As illustrated in Figure \ref{fig:Average_CDS}, $\text{awCDS}$ spikes exponentially specifically for outlier tokens across all architectures.
This structural correlation serves two purposes. 
First, it shows that the spatial instability observed in Table \ref{tab:emd_comparison} is a direct manifestation of outlier tokens being densely composed of highly context-sensitive representations.
Second, together with the CDS distributions in Figure \ref{fig:vit_cis_result}, it further supports the interpretation that CDS isolates feature groups with systematically different degrees of contextual dependency across architectures.

\begin{figure}[htbp]
    \centering
    \captionof{table}{\textbf{Contextual instability of attention maps for outlier vs. non-outlier tokens.} Average EMD scores are computed on the designated layer's attention maps using our SCC method ($s=1$). Across all models, outlier tokens exhibit significantly higher EMD scores, demonstrating severe sensitivity to contextual shifts.} % Table 캡션 추가
    \label{tab:emd_comparison}
    \small % resizebox 대신 폰트 사이즈 조절 사용
    % \begin{tabular}{lcc} % lcc로 수정 (모델명 좌측 정렬)
    \begin{tabular*}{\linewidth}{@{\extracolsep{\fill}} l c c @{}}
    \toprule
    % \multirow{2}{*}{Model} & \multicolumn{2}{c}{EMD} \\ \cmidrule(lr){2-3}
    %                        & Non-outlier ($\bar{D}_{\text{non}}$) & Outlier ($\bar{D}_{\text{out}}$)   \\ \midrule
    Model        & \begin{tabular}[c]{@{}c@{}}non-outlier\\ ($\bar{D}_{\text{non}}$)\end{tabular} & \begin{tabular}[c]{@{}c@{}}outlier\\ ($\bar{D}_{\text{out}}$)\end{tabular} \\ \hline
    DINOv2                 & 0.61        & 6.96      \\ 
    DeiT3                  & 0.38        & 5.97      \\ 
    SigLIP2                & 3.75        & 7.18      \\ \bottomrule
    \end{tabular*}
\end{figure}

\subsection{Feature-Group Removal Analysis Across ViTs}
Having established in \ref{sec:instability_analysis} that other ViT families also exhibit a meaningful CDS-based feature separation, and that outlier tokens remain spatially and representationally unstable across architectures, we next ask whether the downstream functional roles of low-CDS and high-CDS features also generalize beyond CLIP.
To this end, we repeat the feature-group removal analysis of Section \ref{sec:ablation} on DINOv2, SigLIP2, and DeiT3, using the same CDS-based partitioning procedure described in Section \ref{sec:partition}. 
For ImageNet-1K classification, we report results across the three model scales described in Appendix \ref{appendix:sae_other_vit}. 
For ADE20K semantic segmentation and NYUd monocular depth estimation, we use the Base-scale models to keep the dense-prediction setting controlled across architectures.

\begin{figure*}[htbp]
    \centering
    % 0.8\linewidth 또는 0.9\linewidth 등 원하는 비율로 조절
    % 1.0\linewidth로 꽉 채우면 답답해 보일 수 있으니 0.8~0.9가 적절
    \includegraphics[width=1\linewidth]{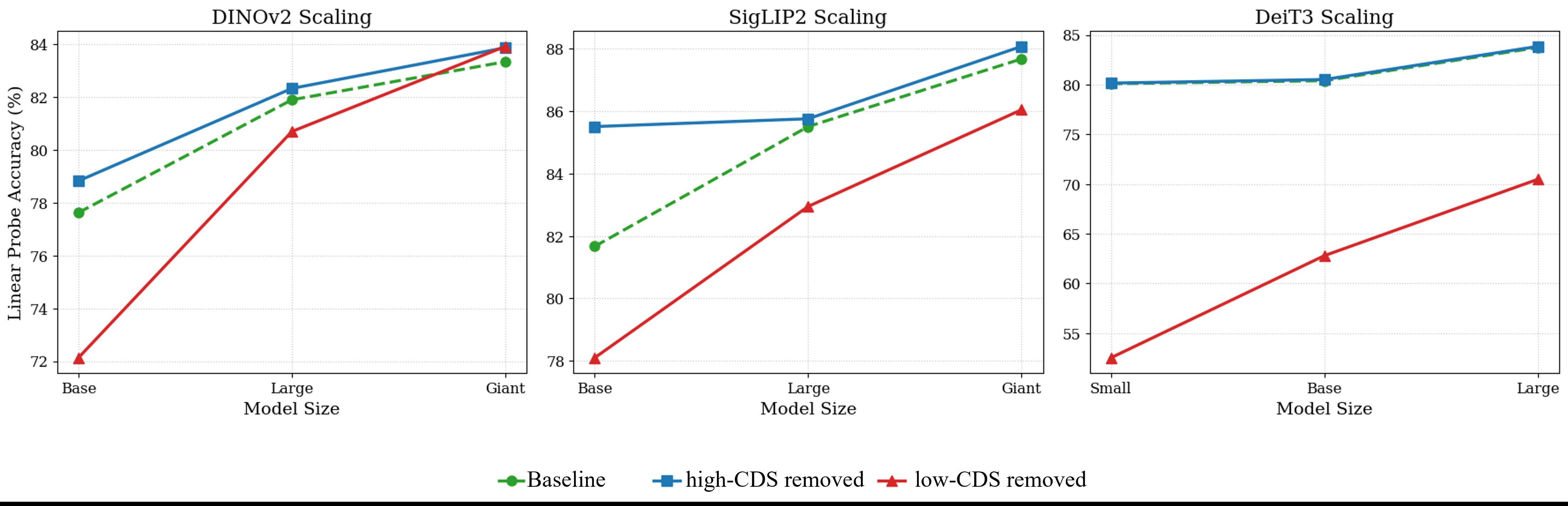} 
    
    % \caption은 figure*에 직접 답니다.
    \caption{\textbf{Linear probe accuracy across ViT scales on ImageNet-1K.} The high-CDS-removed embedding (blue) consistently outperforms the Baseline (green) across all architectures, indicating that local signals are more discriminative for classification. Conversely, the performance gap of the low-CDS-removed embedding (red) generally narrows as model size increases, suggesting that larger models encode broader context more effectively.}
    \label{fig:classification} % 레이블도 의미있게 변경
\end{figure*}

\textbf{Analysis of Linear Probing on Image Classification.}
Figure \ref{fig:classification} summarizes the scaling behavior across the three ViT families. 
Across all three architectures, removing the high-CDS set consistently improves linear-probe accuracy over the original embedding at every scale, while removing the low-CDS set causes a clear performance drop that becomes progressively smaller as model size increases. 
These results extend the trend observed in CLIP: low-CDS features provide the signals that are most directly exploitable by a linear classifier, while the contribution of high-CDS features becomes more competitive as model scale increases.
The weaker gap between the original and low-CDS-removed embeddings in larger models suggests that larger models encode broader contextual information in a form that is increasingly useful for classification. 

\textbf{Analysis of Linear Probing on Dense Prediction Tasks.}
Table \ref{table:ablation_vit} reports the dense-prediction results on ADE20K and NYUd. 
Across all three architectures, removing the low-CDS set causes a substantial degradation on both tasks, confirming that low-CDS features carry the localized information required for dense prediction beyond CLIP. 
In contrast, removing the high-CDS set leaves performance close to the original embedding, with only minor architecture-dependent changes. 
On ADE20K, high-CDS removal improves performance for DINOv2 while causing only slight drops for DeiT3 and SigLIP2. 
On NYUd, its effect is likewise modest but mixed, slightly degrading performance for DINOv2 while improving it for DeiT3 and SigLIP2. 
This consistent asymmetry indicates that high-CDS features provide complementary broader context, whereas low-CDS features remain the primary source of the fine-grained spatial information needed for dense prediction. 
Taken together with the instability results in Table \ref{tab:emd_comparison} and Figure \ref{fig:Average_CDS}, these findings show that the CDS-based functional distinction is not specific to CLIP, but recurs across self-supervised, vision-language, and supervised ViTs.

\begin{table}[]
\caption{\textbf{Linear probing performance of ablated feature sets on other ViT architectures.} We report linear probing results (ADE20K mIoU, NYUd RMSE) for the Base-scale DINOv2, DeiT3, and SigLIP2 models. We compare the original embedding against two ablations: removing low-CDS features and removing high-CDS features. \textbf{BOLD} denotes the best result, while \underline{UNDERLINE} denotes the worst result.}
\resizebox{\columnwidth}{!}{%
\begin{tabular}{llcc}
\hline
Model               & Embedding Type                                                                    & \multicolumn{1}{c}{\begin{tabular}[c]{@{}c@{}}ADE20K\\ mIoU↑\end{tabular}} & \multicolumn{1}{c}{\begin{tabular}[c]{@{}c@{}}NYUd\\ rmse↓\end{tabular}} \\ \hline
DINOv2       & \begin{tabular}[c]{@{}l@{}}Original\\ high-CDS-removed\\ low-CDS-removed\end{tabular} 
& \begin{tabular}[c]{@{}l@{}}\textbf{25.26}\\ 26.11\\ \underline{15.47}\end{tabular}              
& \begin{tabular}[c]{@{}l@{}}\textbf{0.7500}\\ 0.7539\\ \underline{0.8402}\end{tabular}                                                                         \\ \hline
DeiT3       & \begin{tabular}[c]{@{}l@{}}Original\\ high-CDS-removed\\ low-CDS-removed\end{tabular} 
& \begin{tabular}[c]{@{}l@{}}\textbf{26.48}\\ 26.26\\ \underline{8.38}\end{tabular}              
& \begin{tabular}[c]{@{}l@{}}0.8444\\ \textbf{0.8391}\\ \underline{1.0607}\end{tabular}                                                                         \\ \hline
SigLIP2 & \begin{tabular}[c]{@{}l@{}}Original\\ high-CDS-removed\\ low-CDS-removed\end{tabular} 
& \begin{tabular}[c]{@{}l@{}}\textbf{35.62}\\ 35.39\\ \underline{16.93}\end{tabular}              
& \begin{tabular}[c]{@{}l@{}}0.8178\\ \textbf{0.8123}\\ \underline{0.9205}\end{tabular}                                                                         \\ \hline
\end{tabular}%
}
\label{table:ablation_vit}
\end{table}

\begin{table}[]
\caption{\textbf{Sensitivity of non-outlier and outlier tokens under varying shifting factors in SCC.} Outlier tokens consistently exhibit higher instability than non-outlier tokens across all values of $s$, the shifting factor in Shifted Context Crop (SCC).}
\label{tab:scc_ablation}
\resizebox{\columnwidth}{!}{%
\begin{tabular}{lcccccc}
\hline
\multicolumn{1}{c}{\begin{tabular}[c]{@{}c@{}}Shifting\\ factor $s$\end{tabular}} & 1      & 2      & 3      & 4      & 5      & 6      \\ \hline
Non-outlier & 1.490 & 1.745 & 1.842 & 1.855 & 1.915 & 1.958 \\
Outlier     & 5.179 & 6.752 & 6.869 & 6.876 & 5.811 & 5.533 \\ \hline
\end{tabular}%
}
\end{table}

\section{Outlier Tokens' Sensitivity to SCC}
\label{sec:scc_sensitivity}

In Section \ref{sec:pcp_experiment}, we used the Shifted Context Crop (SCC) method to empirically quantify the sensitivity of outlier tokens to contextual shifts. While the main experiments used a fixed shifting factor of $s=1$ to maximize spatial overlap between paired images, it is important to verify whether the observed instability is robust to the magnitude of the spatial shift. To this end, we conducted an ablation study varying $s$ from 1 to 6 using CLIP-L/14-336px.

\subsection{Experimental Setup}
We followed the exact procedure described in Algorithm \ref{alg:PCP} and Section \ref{sec:PCP}. The shifting factor $s$ determines the displacement magnitude between the two generated views $I_1$ and $I_2$. A larger $s$ implies a greater difference in absolute positional embeddings for corresponding patches and a more substantial alteration in global context due to the reduced overlap. We measured the average Earth Mover's Distance (EMD) for both non-outlier and outlier tokens across all heads in the target layer.

\subsection{Results and Analysis}
The results are summarized in Table \ref{tab:scc_ablation}. We observe two distinct trends. The EMD scores for non-outlier tokens exhibit a monotonic increase as $s$ grows, from 1.490 at $s=1$ to 1.958 at $s=6$. This is expected: as the spatial shift increases, the discrepancy in positional information and the divergence of the surrounding context become more pronounced, leading to naturally larger variations in attention patterns.
However, the instability of outlier tokens remains consistently high across all values of $s$. Notably, the EMD scores for outlier tokens are substantially larger than those of non-outlier tokens under every setting. The score peaks around $s=3$ and $s=4$, then decreases slightly at larger shifts. This mild decline at higher shifts may be due to the reduced size of the overlapping region, which limits the effective context available for the attention mechanism to form high-confidence outliers.

This experiment confirms that contextual instability is an intrinsic property of outlier tokens rather than an artifact of a particular shifting factor. Even at the minimal shift of $s=1$, the disparity between non-outlier and outlier tokens remains stark. We therefore adopted $s=1$ in the main experiments to preserve the largest possible evaluation area while still exposing the instability of outlier tokens.

\begin{figure*}[htbp]
    \centering
    \includegraphics[width=0.9\linewidth]{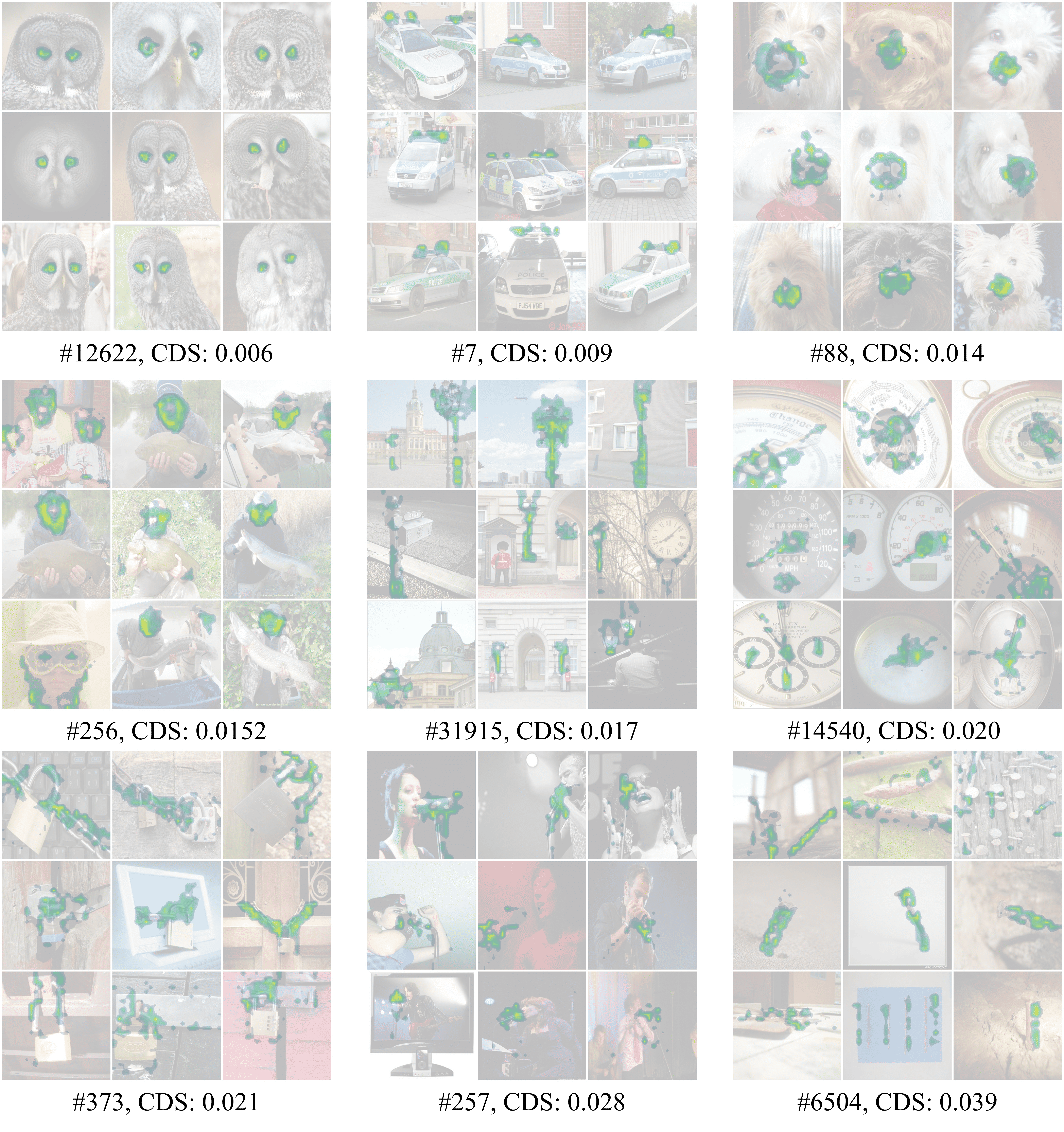} 
    \caption{\textbf{Visualization of low-CDS features.} These features exhibit strong spatial grounding, consistently localizing specific visual concepts across different images.}
    \label{fig:low_cds_visual}
\end{figure*}

\begin{figure*}[htbp]
    \centering
    \includegraphics[width=0.9\linewidth]{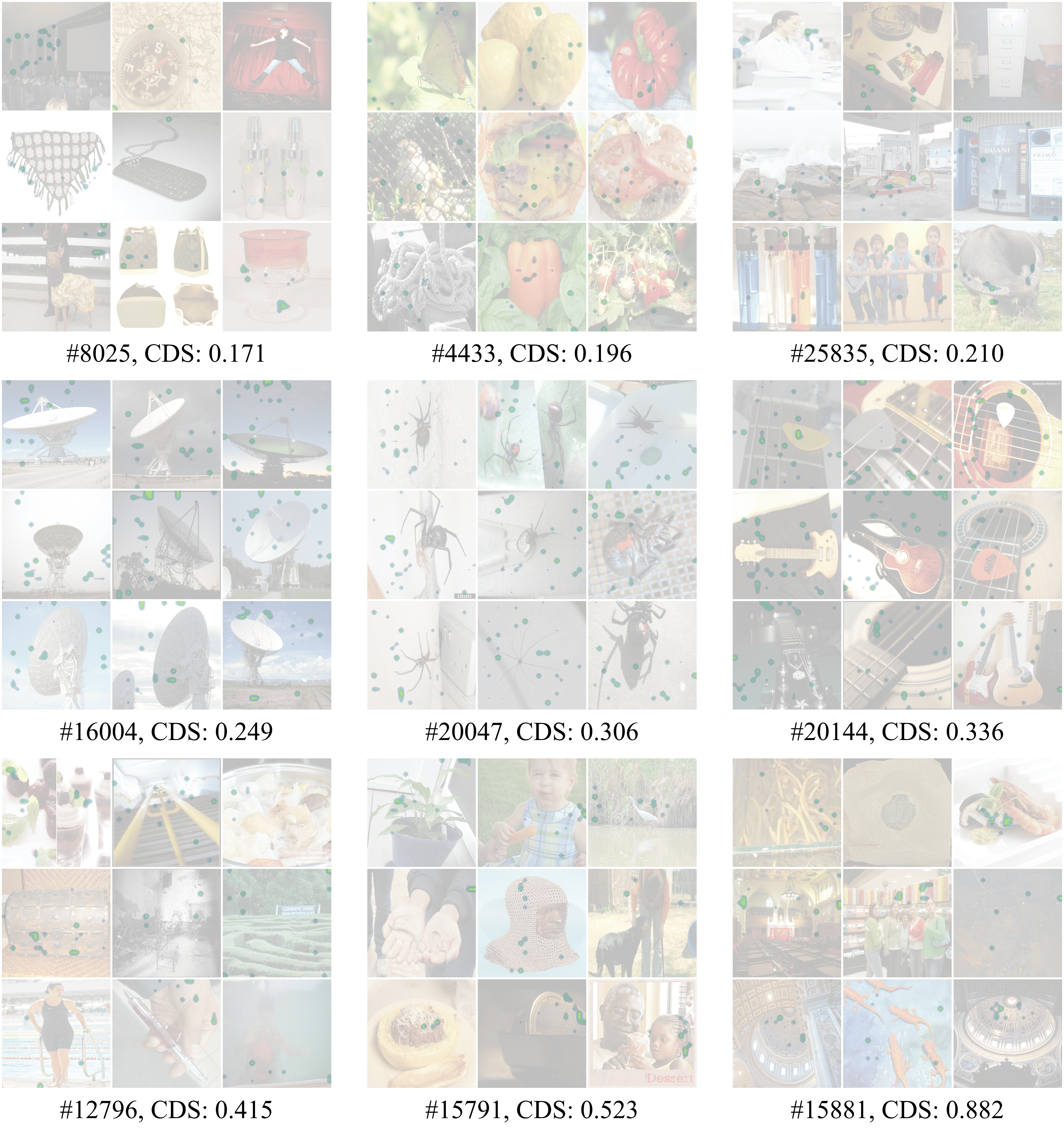} 
    \caption{\textbf{Visualization of high-CDS features.} These features exhibit diffuse activation patterns and capture broader contextual information rather than localized visual details.}
    \label{fig:high_cds_visual}
\end{figure*}

\end{document}